\documentclass[12pt]{iopart}
\usepackage[utf8]{inputenc}
\usepackage{graphicx}
\usepackage{appendix}
\usepackage{caption} 
\usepackage{courier}
\usepackage{graphicx}
\usepackage{amstext}
\usepackage{amssymb}
\usepackage{amsfonts}
\usepackage{booktabs}
\usepackage{array}
\usepackage{arydshln}
\usepackage{soul}
\usepackage{lineno}
\usepackage{color}
\usepackage{multirow}
\usepackage[table,xcdraw]{xcolor}
\usepackage[normalem]{ulem}
\useunder{\uline}{\ul}{}

\usepackage[comma, authoryear]{natbib}
\usepackage{bm}
\usepackage{iopams}

\usepackage[colorlinks=false,hypertexnames=false]{hyperref}
\usepackage{hypcap}
\usepackage{multirow,makecell,varwidth}
\usepackage[hyphenbreaks]{breakurl}
\usepackage{lipsum}

\usepackage{array}
\usepackage{longtable}
\usepackage{lscape}

\hypersetup{
    colorlinks=true,
    linkcolor=red,
    filecolor=blue,      
    urlcolor=blue,
    citecolor=blue,
}
 
\begin{document}

\title[Author guidelines for IOP Publishing journals in  \LaTeXe]{Towards Label-efficient Automatic Diagnosis and Analysis: A Comprehensive Survey of Advanced Deep Learning-based Weakly-supervised, Semi-supervised and Self-supervised Techniques in Histopathological Image Analysis}

\author{Linhao Qu, Siyu Liu, Xiaoyu Liu, Manning Wang\footnote{Corresponding Author.}, Zhijian Song\footnotemark[2]}

\address{Digital Medical Research Center, School of Basic Medical Science, Fudan University, Shanghai Key Lab of Medical Image Computing and Computer Assisted Intervention, Shanghai 200032, China}
\ead{\{lhqu20, mnwang, zjsong\}@fudan.edu.cn.}
\vspace{10pt}

\begin{abstract}
  Histopathological images contain abundant phenotypic information and pathological patterns, which are the gold standards for disease diagnosis and essential for the prediction of patient prognosis and treatment outcome. In recent years, computer-automated analysis techniques for histopathological images have been urgently required in clinical practice, and deep learning methods represented by convolutional neural networks have gradually become the mainstream in the field of digital pathology. However, obtaining large numbers of fine-grained annotated data in this field is a very expensive and difficult task, which hinders the further development of traditional supervised algorithms based on large numbers of annotated data. More recent studies have started to liberate from the traditional supervised paradigm, and the most representative ones are the studies on weakly supervised learning paradigm based on weak annotation, semi-supervised learning paradigm based on limited annotation, and self-supervised learning paradigm based on pathological image representation learning. These new methods have led a new wave of automatic pathological image diagnosis and analysis targeted at annotation efficiency. With a survey of over 130 papers, we present a comprehensive and systematic review of the latest studies on weakly supervised learning, semi-supervised learning, and self-supervised learning in the field of computational pathology from both technical and methodological perspectives. Finally, we present the key challenges and future trends for these techniques.
\end{abstract}


\section{Introduction}
\label{sec1}
Histopathological images contain abundant phenotypic information and pathological patterns, which are the gold standards for disease diagnosis and essential for the prediction of patient prognosis and treatment outcome (Myronenko \emph{et al.}~\citeyear{12}, Wang \emph{et al.}~\citeyear{13}, Srinidhi \emph{et al.}~\citeyear{42}). For clinical diagnosis, experienced pathologists usually require exhaustive examination and interpretation of hematoxylin-eosin-stained (H\&E) tissue slides under a high magnification microscope, including differentiation of tumor areas from large areas of normal tissues, elaborate grading of tumors, and detailed assessment of tumor progression and invasion (e.g., presence of invasive carcinoma or proliferative changes, etc.). This is a highly time-consuming and labor-intensive task, and for example, it usually takes an experienced histopathologist 15 to 30 minutes to examine a complete slide (Wang \emph{et al.}~\citeyear{13}). Moreover, even an experienced pathologist may not be able to accurately determine the deep features hidden in the pathological images, such as predicting lymph node metastasis and prognosis from the primary lesion. Therefore, computer-assisted automatic analysis techniques for histopathological images are in urgent need in clinical practice.

With the advent and development of digital slide scanners in the past two decades, tissues on biopsies can be converted into digital whole slide images (WSIs) that fully preserve the original tissue structure, laying the foundation for automatic pathological image analysis. Early studies in the field of digital pathology diagnosis primarily focused on extracting hand-crafted features from manually selected regions of interest (ROI) by pathologists (Jafari \emph{et al.}~\citeyear{14}, Basavanhally \emph{et al.}~\citeyear{15}, Mercan \emph{et al.}~\citeyear{25}, Yu \emph{et al.}~\citeyear{16}, Luo \emph{et al.}~\citeyear{17}, Qaiser \emph{et al.}~\citeyear{18}) and using machine learning methods (Doyle \emph{et al.}~\citeyear{19}, Rajpoot \emph{et al.}~\citeyear{20}, Qureshi \emph{et al.}~\citeyear{21}, Doyle \emph{et al.}~\citeyear{22}) for automatic analysis and diagnosis. In this regard, Gurcan \emph{et al.}~\citeyear{23} and Madabhushi \emph{et al.}~\citeyear{24} have presented an elaborate review.

In recent years, thanks to the powerful and automatic feature extraction capability, deep learning methods represented by Convolutional Neural Network (CNN) have gradually become the mainstream in the field of digital pathology. However, a major challenge is the huge size of WSIs, typically reaching 100000$\times$100000 pixels at the highest resolution, which prevents the direct use of the entire WSIs as the input to deep learning models. Therefore, when using CNNs to process pathological images, WSIs are usually tiled into many small patches to reduce the computational burden. Earlier studies usually adopted a strongly supervised approach based on these patches to train the network and perform the corresponding classification (Cruz-Roa \emph{et al.}~\citeyear{26}, Cruz-Roa \emph{et al.}~\citeyear{27}, Wei \emph{et al.}~\citeyear{28}, Ehteshami \emph{et al.}~\citeyear{29}, Nagpal \emph{et al.}~\citeyear{30}, Shaban \emph{et al.}~\citeyear{31}, Halicek \emph{et al.}~\citeyear{32}) and segmentation tasks (Chen \emph{et al.}~\citeyear{37}, Gu \emph{et al.}~\citeyear{38}, Swiderska \emph{et al.}~\citeyear{39}). In these works, detailed patch-level annotation is essential, e.g., supervised classification problems require pathologists to give detailed class labels for each patch, and segmentation problems require pathologists to give more detailed pixel-level annotation for each patch.

Although supervised deep learning methods have achieved unprecedented success in digital pathology, they share a common drawback: they all require large amounts of high-quality fine-grained labeled data (patch-level labeled data for classification problems or pixel-level labeled data for segmentation problems) for training. Unfortunately, in the field of digital pathology, obtaining a large amount of data with fine-grained annotation is a very expensive and challenging task, mainly because 1) only experienced pathologists can perform the annotation, and these pathologists are scarce; 2) histopathological images often contain complex and diverse instances of objects, resulting in a large amount of time-consuming and laborious manual annotation effort (Tajbakhsh \emph{et al.}~\citeyear{40}, Yang \emph{et al.}~\citeyear{41}, Srinidhi \emph{et al.}~\citeyear{42}). Arguably, the lack of a large amount of annotated data limits the application of deep learning techniques in computational pathology. For this reason, some new studies have recently attempted to liberate from the traditional strongly supervised paradigms, the most representative of which are the weakly supervised learning paradigm based on weak annotations, the semi-supervised learning paradigm based on limited annotations, and the self-supervised paradigm based on the representation learning of pathological images.

The weakly supervised learning paradigm no longer requires pathologists to give annotations of all pixels or regions on the entire WSI, but only class labels or sparse region annotations on the entire WSI; the semi-supervised learning paradigm no longer requires pathologists to give fine-grained annotations of a large amount of data, but only a small fraction of fine-grained labeled data and a large amount of unlabeled data; while the self-supervised learning paradigm can create supervised information through a large amount of unlabeled data for self-supervised training to learn an accurate feature representation of the data. In the process of training with limited labeled data, using the features trained by self-supervised learning to determine the initial model weights can significantly improve the performance of the model. Therefore, weakly supervised learning, semi-supervised learning and self-supervised learning are leading a new study direction of the automatic diagnosis and analysis for pathological images.

However, there are very few related reviews. Srinidhi \emph{et al.}~\citeyear{42} reviewed representative supervised learning, weakly supervised learning, unsupervised learning, and transfer learning studies in the field of computational pathology until December 2019. Rony \emph{et al.}~\citeyear{43} reviewed representative weakly supervised learning studies until 2020. Nevertheless, in recent years, deep learning techniques have been developing rapidly and the new techniques continue to emerge. Therefore, a review regarding the applications of these techniques in the automatic diagnosis of pathological images has important theoretical value and clinical significance.

In this review, we summarize more than 130 recent technical studies systematically on weakly supervised learning, semi-supervised learning, and self-supervised learning in the field of computational pathology. We performed this extensive review by searching Google Scholar, PubMed, and arXiv for papers including keywords such as ("deep learning" or "weakly supervised learning" or "semi-supervised learning" or "self-supervised learning ") and ("digital pathology" or "histopathology" or "computational pathology"). Notably, on the one hand, we focus on papers presenting novel techniques and theories with high impact (h-index, citations and impact factors of journals), thus we concentrate more on studies published in top conferences (including CVPR, NeurIPS, MICCAI, ISBI, MIDL, IPMI, AAAI, ICCV, ECCV, etc.) and top journals (including TPAMI, TMI, MIA, etc.) on weakly supervised, semi-supervised, and self-supervised learning in the field of computational pathology. On the other hand, since technical research in this area is growing rapidly and more new techniques have been proposed, we mainly cover papers published in 2019-2021. On the other hand, we also present a meticulous summary of the disease types, tasks, datasets, and performance covered by these papers. In total, this review contains more than 200 relevant references.

The rest of the paper is organized as follows: Section \ref{sec2} expounds a general overview of the weakly supervised, semi-supervised, and self-supervised learning paradigms in the context of computational pathology; Section \ref{sec3} includes a detailed review of the weakly supervised (Section \ref{sec31}), semi-supervised (Section \ref{sec32}), and self-supervised (Section \ref{sec33}) learning paradigms; We discuss the three learning paradigms and their future trends in Section \ref{sec4}, and conclude the whole paper in Section \ref{sec5}. The list of all the acronyms used in this review is shown in Table \ref{tab1}.

\begin{table}[htbp]
  \tiny
  \centering
  \caption{List of all the acronyms in this review.}
  \begin{tabular}{llll}
  \hline
  \textbf{Full   Name}                                            & \textbf{Acronyms} & \textbf{Full Name}                            & \textbf{Acronyms} \\ \hline
  Area Under ROC Curve & AUC               & Graph Neural Network                          & GNN               \\
  Auxiliary Classier Generative   Adversarial Networks            & AC-GAN            & Hematoxylin-Eosin-Stained                     & H\&E              \\
  Average Hausdorff Distance                                      & AHD               & Magnication Prior Contrastive Similarity      & MPCS              \\
  Average Jaccard Index                                           & AJI               & Mean Average Precision                        & MAP               \\
  Calinski-Harabaz Index                                          & CHI               & Mean Teachers                                 & MT                \\
  Contrastive Predictive Coding                                   & CPC               & Microsatellite Instability                    & MSI               \\
  Convolutional Autoencoder                                       & CAE               & Multiple Instance Fully Convolutional Network & MI-FCN            \\
  Convolutional Neural Network                                    & CNN               & Multiple Instance Learning                    & MIL               \\
  Deep Learning Hashing                                           & DLH               & Noise Contrastive Estimation                  & NCE               \\
  Deformation Representation   Learning                           & DRL               & Percentage Of Tumor Cellularity               & TC                \\
  Diffusion-Convolutional Neural   Networks                       & DCNNs             & Recurrent Neural Network                      & RNN               \\
  Dual-Stream Multiple Instance   Learning                        & DSMIL             & Regions Of Interest                           & ROI               \\
  Expectation-Maximization                                        & EM                & Resolution Sequence Prediction                & RSP               \\
  Exponential Moving   Average                                    & EMA               & Silhouette Index                              & SI                \\
  Focal-Aware Module                                              & FAM               & Support Vector Machines                       & SVM               \\
  Frechet Inception Distance                                      & FID               & Temporal Ensembling                           & TE                \\
  Generative Adversarial   Networks                               & GAN               & The Cancer Genome Atlas Program               & TCGA              \\
  Graph Convolutional Neural   Network                            & GCN               & Whole Slide Images                            & WSI               \\ \hline
  \end{tabular}
  \label{tab1}
  \end{table}

\section{Overview of Learning Paradigms and Problem Formulation}
\label{sec2}
\begin{figure*}[htbp]
    \centering
    \includegraphics[scale=0.23]{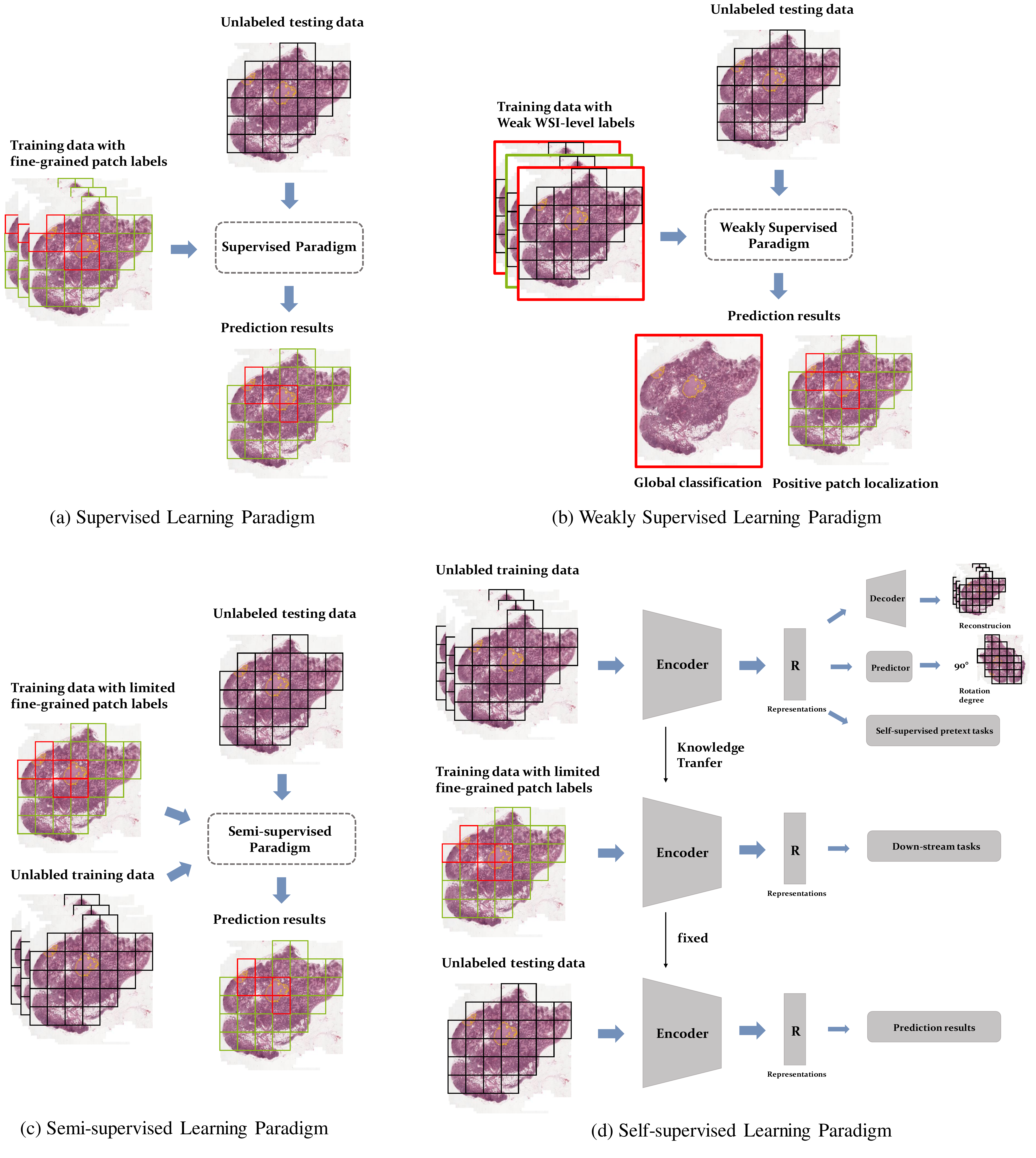}
    \caption{General overview of the learning paradigms reviewed in this paper, depicted as an example of classifying normal tissue (green) and cancerous tissue (red) in a WSI. Note that the training data and testing data in this figure are used for description only and are not necessarily the real case. (a) Supervised learning paradigm. (b) Weakly Supervised learning paradigm. (c) Semi-supervised learning paradigm. (d) Self-supervised learning paradigm.}
    \label{fig1}
\end{figure*}

In this section, we provide a general overview and problem formulation of the three learning paradigms reviewed in this paper, and compare them with the traditional strongly supervised paradigm. To make the description more specific and vivid, we present an example of accurately classifying normal and cancerous tissues in a WSI, as shown in Figure \ref{fig1}. The raw data for this example WSI comes from a study on predicting lymph node metastasis in breast cancer using deep learning (Bejnordi \emph{et al.}~\citeyear{44}). We also intuitively compare and summarize these paradigms in Table \ref{tab2}.

For the dataset $W={\{W_i\}}_{i=1}^N$ consisting of $N$ WSIs, each WSI $W_i$ is now cut into patches $\{p_{i,j},j=1,2,\ldots n_i\}$, and $n_i$ is the number of patches cut out of $W_i$. In the supervised learning paradigm, a large number of patches with fine-grained labels are available for training, so each patch is given a label $y_{i,j}\in\mathbb{R}^{C}$, and $C$ denotes the possible class. For example, in the binary classification task, $C=2$ and the label takes the scalar form \{0, 1\} while in the regression task, $C$ takes the form of a continuous set of real numbers $\mathbb{R}$. The goal of the supervised learning paradigm is to train a model $f_\theta:x\rightarrow y$ to optimally predict the labels $y_{i,j}$ of the unknown patches $p_{i,j}$ in the test WSI based on the loss function $\mathcal{L}$. Figure \ref{fig1} (a) illustrates the main process of this paradigm. During training, the model is trained in a supervised manner using patches cut out of the training WSIs and their labels (green for negative and red for positive) by pathologists; during testing, the trained model is used to predict the labels of the patches cut out of the unseen test WSIs.

In the weakly supervised learning paradigm, the label $y_{i,j}$ of each patch is typically unknown, while only the label of each WSI is available, and thus the traditional strongly supervised learning paradigm cannot work. In this review, we focus on the most dominant weakly supervised paradigm currently used in computational pathology, the deep multiple instance learning (MIL) approach. In MIL, each WSI is considered as a bag containing many patches (also called instances). if a WSI (bag) is labeled as disease-positive, then at least one patch (instance) in that WSI is disease-positive; if a WSI is disease-negative, then all patches in that WSI are negative. The relationship between a WSI (bag) and its patches (instances) can be expressed mathematically as follows.

Given a dataset $W={\{W_i\}}_{i=1}^N$ consisting of $N$ WSIs, each image $W_i$ has a corresponding label $Y_i\in\left\{0,1\right\},\ i=\{1,2,...N\}$. Now each WSI $W_i$ is cut into small patches $\{p_{i,j},j=1,2,\ldots,n_i\}$ without overlapping each other, and $n_i$ is the number of patches. All patches $\{p_{i,j},j=1,2,\ldots, n_i\}$ in $W_i$ form a bag, the bag-level label is the label $Y_i$ of $W_i$, and each small patch is called an instance of this bag, while the instance-level label $y_{i,j}$ and its corresponding bag-level label $Y_i$ have the following relationship:
\begin{equation}
    Y_{i}=\left\{\begin{array}{cc}
        & 0, \text { if } \sum_{j} y_{i, j}=0 \\
        & 1, \text { else }
        \end{array}\right.  \label{eq1}
\end{equation}

It means that the labels of all instances in the negative bag are negative, while at least one positive instance exists in the positive bag and the labels of instances $y_{i,j}$ are unknown.

As shown in Figure \ref{fig1} (b), generally, there are two main goals of deep learning-based WSI analysis, one is global slide classification, i.e., to accurately classify each WSI, and the other is positive patch localization, i.e., to accurately classify each instance in positive bags. A review of the current state-of-the-art weakly supervised learning methods is presented in Section \ref{sec31}.

In the semi-supervised learning paradigm, we only have a very small number of patches with labels, in addition to a large number of unlabeled patches that can also be used for training. Therefore, the main goal of the semi-supervised learning paradigm is how to use the unlabeled data to improve the performance of the models trained with limited labeled data. As shown in Figure \ref{fig1} (c), in contrast to the supervised learning paradigm, the semi-supervised learning paradigm makes use of a large amount of unlabeled data while training with the labeled data. During testing, the trained model is used to predict the labels of the patches in test WSIs. See Section \ref{sec32} for a detailed review of the semi-supervised learning methods.

Self-supervised learning is a hybrid learning approach, which combines unsupervised and supervised learning paradigms in a pre-training and fine-tuning manner. The aim is to get better results of supervised training though generating supervised information from a large amount of unlabeled data, which can learn better feature representations, and can reduce manual annotation in the subsequent tasks. Due to the small amount of annotated data, it is not sufficient to use these data directly to train the model. Therefore, the self-supervised learning paradigm first learns a primary feature representation from a large amount of unlabeled data, which is called the pre-training process. The feature representations learned in the self-supervised auxiliary tasks are then transferred for further training in downstream tasks using limited labeled data, which is called the fine-tuning process. In this way, the primary feature representations can effectively help the network to achieve an effective training result with less labeled data.

As shown in Figure \ref{fig1} (d), the pre-training process of the self-supervised learning paradigm is typically performed through self-supervised auxiliary tasks. In the self-supervised auxiliary tasks, certain inherent properties of the unlabeled data are first utilized to generate supervised information, and then the network is trained by the self-supervised information, such as self-reconstruction, random rotation followed by angle prediction, color information discarding followed by colorization, and patch position disruption followed by restoration. Once accomplishing these self-supervised auxiliary tasks, the effective feature representations can be extracted. The fine-tuning process of self-supervised learning is done in the downstream tasks. During the fine-tuning process, a small amount of labeled data is used to perform the supervised training, and the model is not trained from scratch, but is further trained using the feature representations learned in the auxiliary tasks as the initial weights of the network. Finally, the trained network is used for testing. A review of the state-of-the-art self-supervised learning methods is presented in Section \ref{sec33}.

\begin{table}[!t]
  \tiny
  \centering
  \caption{Intuitive summary and comparison of the four paradigms.}
  \begin{tabular}{m{1.5cm}<{\centering}m{1.8cm}<{\centering}m{3cm}<{\centering}m{2.8cm}<{\centering}m{2cm}<{\centering}m{1.5cm}<{\centering}}
  \hline
  \textbf{Methods}                      & \textbf{Input}                                                                 & \textbf{Suitable tasks}                                                                       & \textbf{Technical paradigms}                                                                                                                                    & \textbf{Strengths}                                                                                                                       & \textbf{Weaknesses}                                                                                 \\ \hline
  Supervised learning paradigm        & A large number of small patches (tiled from WSIs) with fine-grained labels   & WSI-level and patch-level classification/segmentation/regression                            & \multicolumn{1}{c}{-}                                                                                                                                           & Broad application, effective and simple training                                                                                       & Require large amount of fine-grained labeled data                                                 \\ \hline
  Weakly Supervised learning paradigm & Entire WSIs with overall labels or sparse labels                             & WSI-level classification/segmentation/regression, Patch-level coarse-grained localization & Instance-based approach, Bag-based approach, Hybrid approach                                                                                                  & No need for fine-grained annotation, effectively reduce the burden of data annotation                                                  & Achieve limited performance for fine-grained tasks                                                \\ \hline
  Semi-supervised learning paradigm   & A limited number of small patches (tiled from WSIs) with fine-grained labels & WSI-level and patch-level classification/segmentation/regression                            & Pseudo-labelling-based approach, Consistency-based approach, Graph-based approach, Unsupervised-preprocessing-based approach, GAN-based approach and others & Require only a small amount of fine-grained annotation, effectively reduce the burden of data annotation                               & Need to satisfy various semi-supervised assumptions                                               \\ \hline
  Self-supervised learning paradigm   & A large number of small patches (tiled from WSIs) without labels             & Patch-level feature representations, Multiple related down-stream tasks                     & Predictive approach, Generative approach, Contrastive approach, Hybrid approach                                                                               & Efficiently extract image features from a large amount of unsupervised data, effectively reduce the data annotation burden & May result in information loss when the extracted features are not applicable to downstream tasks \\ \hline
  \end{tabular}
  \label{tab2}
  \end{table}

\section{Paradigms}
\label{sec3}
\subsection{Weakly Supervised Learning Paradigm}
\label{sec31}
In this section, we provide a comprehensive review of the primary deep multiple instance learning (MIL) methods currently used in the weakly supervised learning paradigm for computational pathology. In MIL, each WSI is considered as a bag containing many patches (also called instances). If a WSI (bag) is labeled disease-positive, then at least one patch (instance) in that WSI is disease-positive; if a WSI is disease-negative, then all patches in that WSI are negative.

We categorize the current deep MIL methods for WSI analysis into instance-based methods, bag-based methods, and hybrid methods. Our categorization is mainly based on whether the methods contain an instance classifier or a bag classifier, i.e., instance-based methods contain only an instance classifier; bag-based methods contain only a bag classifier; while hybrid methods contain both an instance classifier and a bag classifier. In this way, the categories clearly cover almost current deep MIL methods for WSI analysis. A diagram of the three methods above is shown in Figure \ref{fig2}. The detailed literatures in this section are summarized in Table \ref{table1}.

\subsubsection{Instance-based Approach}
\label{sec311}
\begin{figure*}[htbp]
    \centering
    \includegraphics[scale=0.16]{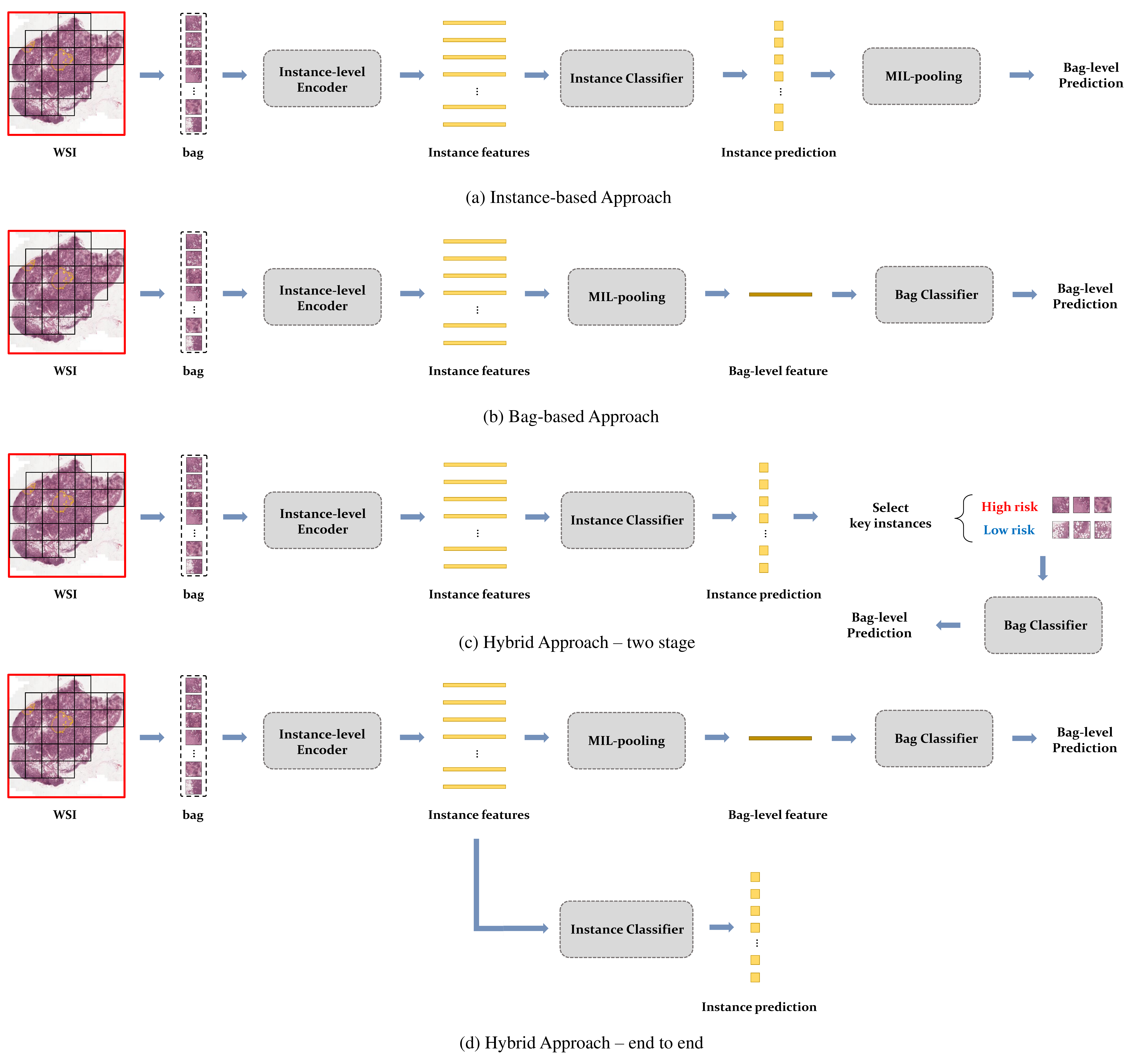}
    \caption{Overview of multiple instance learning methods. (a) Instance-based Approach. (b) Bag-based Approach. (c) Two-stage Hybrid Approach. (c) End-to-End Hybrid Approach.}
    \label{fig2}
\end{figure*}

\
\newline
The main idea of the instance-based approach is to train a good instance classifier to accurately predict the potential labels of instances in each bag, and then use MIL-pooling to aggregate the predictions of all instances in each bag to obtain the prediction of the bag. The details are shown in Figure \ref{fig2} (a). Since the true labels of each instance are unknown, these approaches usually first assign the labels of each instance with their corresponding bags as the pseudo labels (i.e., all instances in a positive bag are given positive labels, and all instances in a negative bag are given negative labels), and then train the instance classifier using a supervised way until it converges. The loss function is usually the cross-entropy function defined between the predictions of the instance classifier and the pseudo labels. After training, the instance classifier is used to make predictions for all instances in the test bag, and then the predictions of each instance are aggregated to obtain the prediction of the bag, and this aggregation process is called MIL-pooling. Commonly used MIL pooling methods include Mean-pooling (Wang \emph{et al.}~\citeyear{3}), Max-pooling (Feng \emph{et al.}~\citeyear{2}, Wang \emph{et al.}~\citeyear{3}, Wu \emph{et al.}~\citeyear{4}), Voting (Cruz-Roa \emph{et al.}~\citeyear{26}), log-sum-exp-pooling (Ramon \emph{et al.}~\citeyear{6}), Noisy-or-pooling (Maron \emph{et al.}~\citeyear{8}), Noisy-and-pooling (Kraus \emph{et al.}~\citeyear{7}), and Dynamic pooling (Yan \emph{et al.}~\citeyear{1}) among others.

Instance-based approach is more common in early studies, and its main advantage lies in the direct prediction of each instance so that the localization task can be performed conveniently. However, it has two major drawbacks. First, since the true labels of each instance in the positive bags are not necessarily all positive, the pseudo labels assigned to the instances in the positive bags are noisy, which will lead to inaccurate training of the instance classifier; Second, the MIL-pooling method, which aggregates the predictions of instances in each bag, is manually designed and non-trainable, making it less flexible and robust. Therefore, the performance of these methods is usually limited.

\subsubsection{Bag-based Approach}
\label{sec312}
\
\newline
The main idea of the bag-based approaches is to first extract the features of each instance in a bag using shared instance-level feature extractors, then use MIL-pooling to aggregate the instance-level features to obtain the bag-level features, and then train the bag classifier in a supervised manner until it converges. The specific diagram is shown in Figure \ref{fig2} (b). The loss function is usually defined as the cross-entropy loss between the predictions of the bag classifier and the true bag labels.

MIL-pooling also exists in bag-based methods, but unlike instance-based methods, MIL-pooling here aggregates not the predictions of instances, but the features of instances. Mean-pooling, Max-pooling and other aggregation methods can also be used as aggregation methods for instance features, but their drawbacks remain, i.e., they cannot be trained and adjusted adaptively, so they are often not flexible enough.

The key of the bag-based methods is the training of the bag classifier. Since the true labels of the bags are available, there is no noise in their training process, so these methods tend to be more accurate than instance-based methods in bag classification. However, a serious problem of the bag-based approaches is that they cannot perform the localization task easily. Furthermore, the aggregation functions for instance features are not flexible enough to show the contribution of different instances to bag classification.

\paragraph{Attention-based Approach}
Ilse \emph{et al.}~\citeyear{176} have alleviated these dilemmas. They first proposed to use the trainable attention mechanism to aggregate instance features, and started a wave of study on attention-based aggregation methods by subsequent bag-based methods. They trained both the instance-level feature extractor and a bag-level classifier using an end-to-end manner, and used the attention mechanism to aggregate the features and measure the significance of each instance. Tu \emph{et al.}~\citeyear{45} proposed a new end-to-end graph neural network (GNN) for instance aggregation. This work is the first GNN-based MIL work. 
Hashimoto \emph{et al.}~\citeyear{11} proposed a novel end-to-end method for cancer subtype classification by combining MIL, domain adversarial and multiscale learning frameworks. Yao, Zhu \emph{et al.}~\citeyear{46,47} proposed a deep attention guided MIL framework for cancer survival analysis. They first used a pre-trained model from ImageNet (Deng \emph{et al.}~\citeyear{168}) to extract the features of instances in each bag, and then used K-means algorithm to cluster the instances in each bag to obtain the phenotypic patterns, and finally applied attention mechanism to aggregate the features of these patterns and performed prediction.

\paragraph{Self-supervised Pre-training-based Approach}
Due to the extremely large size of WSIs and the large number of instances cut out, direct end-to-end training of all instances is easily limited by computational resources. Therefore, some studies first use advanced self-supervised pre-training methods to characterize each instance and then perform subsequent training. Lu \emph{et al.}~\citeyear{98} first proposed to obtain instance-level feature representations by self-supervised contrastive predictive coding (CPC), and then used the attention-based MIL method for instance aggregation to perform bag-level classification. This is the first MIL study using self-supervised contrastive learning. Zhao \emph{et al.}~\citeyear{192} used a pre-trained VAE-GAN (Larsen \emph{et al.}~\citeyear{162}) to extract instance-level features, and then used GNN to aggregate instance features and perform bag-level classification. Li \emph{et al.}~\citeyear{191} proposed DSMIL, where they used contrastive pre-training (Chen \emph{et al.}~\citeyear{142}) to obtain the instance features, and then proposed the masked non-local operation-based dual-stream aggregator to perform both instance-level classification and bag-level classification.

\paragraph{Transformer Based Approach}
In MIL-based WSI analysis, not only the contribution of different instances to bag classification should be considered, the relationships among different instances should also be fully explored, because different instances in a WSI are not isolated from each other, but have strong correlation. To address this issue, Shao \emph{et al.}~\citeyear{48} and Li \emph{et al.}~\citeyear{49} et al. used Transformer-based architectures to aggregate instances and both achieved promising results. The former designed a Transformer-based correlated MIL framework to explore the morphological and spatial information among different instances and provided related proofs. The latter presented a MIL framework based on the deformable transformer and convolutional layers.

\subsubsection{Hybrid Approach}
\label{sec313}
\
\newline
The hybrid approach combines the advantages of the above two approaches. It trains both the instance-level classifier and the bag-level classifier, and uses the former to predict the instance-level results while the latter for bag-level results. Overall, there are two types of the hybrid approaches. One is the two-stage approach and the other is the end-to-end approach.

\paragraph{Two-stage Hybrid Approach}
The two-stage hybrid approach generally trains the instance classifier by assigning each instance in each bag with their corresponding bag labels as pseudo labels, and then trains the bag classifier to complete the bag classification based on the predictions of the instance classifier. Some studies have also attempted to select the key instances in each bag based on the predictions of the instance classifier, and then train the bag classifier based on these key instances. The specific diagram is shown in Figure \ref{fig2} (c).
Hou \emph{et al.}~\citeyear{50} proposed a new Expectation-Maximization (EM) based model. They selected discriminative instances based on spatial relationship to train the instance classifier and fed the histogram of instance predictions into the multiclass logistic regression model and the SVM model (Chang \emph{et al.}~\citeyear{51}) for bag prediction. Campanella \emph{et al.}~\citeyear{193} first selected key instances with the maximum prediction probability of the instance classifier in the current iteration and assigned pseudo labels of the corresponding bag labels to them. Then they fed the features of these key instances into the recurrent neural network (RNN) to perform the aggregation and prediction of the bags. Wang \emph{et al.}~\citeyear{13} selected key instances based on the predictions of positive instance probability and fed their features into the global feature descriptor and used the random forest algorithm to classify the bags. Chen \emph{et al.}~\citeyear{53} proposed a focal-aware module (FAM) and used thumbnails of WSI to automatically estimate the key regions associated with the diagnosis. Then, the instance features at different scales were extracted based on these key regions and aggregated using GNN to perform the bag classification.

\paragraph{End-to-end Hybrid Approach}
The end-to-end hybrid approach generally trains the instance-level classifier and the bag-level classifier at the same time. A common approach is to train the two classifiers simultaneously by assigning each instance the corresponding bag labels as pseudo labels on top of the bag classifier. Some studies also train the instance classifier to select the key instances in an epoch first, and then train the bag classifier after aggregating the instance features. The specific diagram is shown in Figure \ref{fig2} (d).
Shi \emph{et al.}~\citeyear{175} proposed loss-based attention MIL. They added an instance-level loss function weighted by the instance attention scores based on AB-MIL (Ilse \emph{et al.}~\citeyear{176}) as a regularization term to improve the recall of instances and used consistency constraints to smooth the training process to improve the generalization ability. Chikontwe \emph{et al.}~\citeyear{177} combined top-k instance selection, instance-level representation learning, and bag-level representation in an end-to-end framework. Sharma \emph{et al.}~\citeyear{54} also combined instance selection, instance-level representation learning and bag-level representation in an end-to-end framework. Unlike (Chikontwe \emph{et al.}~\citeyear{177}), they proposed to use a clustering-based sampling method to select key instances. Lu \emph{et al.}~\citeyear{55} also proposed a MIL framework based on clustering and attention mechanisms. They selected the instances with the largest and smallest attention scores in the current bag for clustering to enhance the learning of feature space. Myronenko \emph{et al.}~\citeyear{12} proposed a MIL framework combining the Transformer and CNN architectures to compute the interrelationships between instances and aggregate the instances features to accomplish the bag classification. They added the instance loss to assist the optimization process.

\subsubsection{Representative Clinical Studies}
\label{sec314}
\
\newline
A large number of outstanding studies have been dedicated to address significant clinical problems using weakly supervised methods. For example, Coudray \emph{et al.}~\citeyear{58} et al. developed deep learning models for accurate prediction of cancer subtypes and genetic mutations and sparked the whole field of weakly supervised computational pathology. 
Naik \emph{et al.}~\citeyear{10} et al. presented an attention-based deep MIL framework to predict directly estrogen receptor status from H\&E slices. Another typical clinical work comes from Tomita \emph{et al.}~\citeyear{9}, who proposed a grid-based attention network to perform 4-class classification of high-resolution endoscopic esophagus and gastroesophageal junction mucosal biopsy images from 379 patients. Skrede \emph{et al.}~\citeyear{56} developed a multi-scale deep MIL-based model to analyze conventional HE-stained slides and developed a model that can effectively predict the prognosis of patients after colorectal cancer surgery. 
Another gastrointestinal tract oncology study (Kather \emph{et al.}~\citeyear{57}) predicted microsatellite instability (MSI) based on a deep MIL model directly on HE-stained slides. Currently, weakly supervised deep-learning models for digital pathological analysis has been applied in a wide range of cancer types including breast, colorectal, lung, liver, cervical, thyroid, and bladder cancers 
(Coudray \emph{et al.}~\citeyear{58}, Chaudhary \emph{et al.}~\citeyear{60}, Wessels \emph{et al.}~\citeyear{61}, Campanella \emph{et al.}~\citeyear{193}, Anand \emph{et al.}~\citeyear{194}, Yang \emph{et al.}~\citeyear{195}, 
Li \emph{et al.}~\citeyear{196}, Saillard \emph{et al.}~\citeyear{197}, Velmahos \emph{et al.}~\citeyear{198}, Woerl \emph{et al.}~\citeyear{199}).

\begin{table}[!htbp]
\centering
\rotatebox[origin=c]{90}{%
\begin{varwidth}{\textheight}
\caption{List of literatures in the weakly supervised learning section.}
\footnotesize
\tabcolsep=16pt
\resizebox{.95\textwidth}{!}{%
\begin{tabular}{ccccccccc}
\toprule
\textbf{Reference}&\textbf{Approach}&\textbf{Disease Type}&\textbf{Staining}&\textbf{Task}&\textbf{Dataset}&\textbf{Dataset Scale}&\textbf{Dataset Link}&\textbf{Performance}\\
\midrule
\multirow{3}{*}{\cite{1}}&\multirow{3}{*}{Instance-based}&Breast Cancer&\multirow{3}{*}{H\&E}&\multirow{3}{*}{\makecell{Benign and\\malignant classification}}&UCSB breast dataset&58 cases&\cite{204}&Accuracy: 0.927\\
&&\makecell{Diabetes (from eye\\fundus images)}&&&Messidor dataset&1200 cases&\cite{207}&Accuracy: 0.740\\
\cite{7}&Instance-based&Breast Cancer&\makecell{Three channels with\\fluorescent markers for\\ DNA, actin filaments,\\ and b-tubulin}&\makecell{Classification of 12\\ distinct categories}&\makecell{Broad Bioimage Benchmark\\Collection (BBBC021v1) Dataset}&340 cases&\cite{205}&\makecell{Accuracy: 0.958 for full\\image, 0.971 for treatment}\\
\cite{26}&Instance-based&Breast Cancer&H\&E&\makecell{Automatic detection of\\invasive ductal carcinoma\\ tissue regions}&\makecell{Clinical histopathology dataset\\collected from multiple hospitals}&162 cases&inhouse&Accuracy: 0.842\\
\multirow{2}{*}{\cite{176}}&\multirow{2}{*}{Bag-based}&Breast Cancer&\multirow{2}{*}{H\&E}&\multirow{2}{*}{\makecell{Automatic detection\\of cancerous regions}}&Breast cancer dataset&58 cases&\cite{206}&Accuracy: 0.755\\
&&Colon Cancer&&&Colon cancer dataset&100 cases&\cite{215}&Accuracy: 0.904\\
\cite{45}&Bag-based&\makecell{Diabetes (from eye\\fundus images)}&H\&E&\makecell{Diagnosing diabetes from \\weakly labeled retinal images}&Messidor dataset&1200 cases&\cite{207}&Accuracy: 0.742\\
\cite{11}&Bag-based&Malignant Lymphoma&H\&E&\makecell{Classification of malignant\\lymphoma sub-types}&\makecell{Clinical histopathology dataset\\collected from multiple hospitals}&196 cases&inhouse&Accuracy: 0.871\\
\multirow{2}{*}{\cite{47}}&\multirow{2}{*}{Bag-based}&Lung Cancer&\multirow{2}{*}{H\&E}&\multirow{2}{*}{Cancer survival prediction}&\makecell{National Lung Screening\\Trial (NLST) dataset}&387 cases&\cite{208}&AUC: 0.652\\
&&Colorectal Cancer&&&\makecell{Molecular and Cellular\\Oncology (MCO) dataset}&1146 cases&\cite{216}&AUC: 0.7143\\
\cite{98}&Bag-based&Breast Cancer&H\&E&Classification of normal or benign&BACH dataset&400 cases&\cite{209}&Accuracy: 0.95\\
\cite{192}&Bag-based&Colon Adenocarcinoma&H\&E&Prediction of lymph node metastasis&\makecell{The Cancer Genome\\Atlas (TCGA) dataset}&425 cases&\cite{210}&Accuracy: 0.6761\\
\multirow{3}{*}{\cite{191}}&\multirow{3}{*}{Bag-based}&Breast Cancer&\multirow{3}{*}{H\&E}&Detection of lymph node metastases&Camelyon16 dataset&400 cases&\cite{211}&Accuracy: 0.8992\\
&&Lung Cancer&&\makecell{Diagnosis of lung\\cancer subtypes}&\makecell{The Cancer Genome Atlas\\(TCGA) lung cancer dataset}&1054 cases&https://portal.gdc.cancer.gov/&Accuracy: 0.9571\\
\multirow{3}{*}{\cite{48}}&\multirow{3}{*}{Bag-based}&Breast Cancer&\multirow{3}{*}{H\&E}&Detection of lymph node metastases&Camelyon16 dataset&400 cases&\cite{211}&Accuracy: 0.8837\\
&&Lung Cancer&&Diagnosis of cancer subtypes&TCGA-NSCLC dataset&993 cases&https://portal.gdc.cancer.gov/&Accuracy: 0.8835\\
&&Kidney Cancer&&Diagnosis of cancer subtypes&TCGA-RCC dataset&884 cases&https://portal.gdc.cancer.gov/&Accuracy: 0.9466\\
\multirow{2}{*}{\cite{49}}&\multirow{2}{*}{Bag-based}&Breast Cancer&\multirow{2}{*}{H\&E}&Detection of lymph node metastases&BREAST-LNM dataset&3957 cases&inhouse&AUC: 0.7288\\
&&Lung Cancer&&Diagnosis of lung cancer subtypes&CPTAC-LUAD dataset&1065 cases&\cite{213}&AUC: 0.9906\\
\multirow{3}{*}{\cite{50}}&\multirow{3}{*}{Hybrid}&Glioma&\multirow{3}{*}{H\&E}&Classification of glioma&\multirow{3}{*}{\makecell{The Cancer Genome Atlas (TCGA) dataset}}&209cases&\multirow{3}{*}{https://portal.gdc.cancer.gov/}&Accuracy: 0.771\\
&&Lung Cancer&&\makecell{Diagnosis of non-small-cell\\lung carcinoma subtypes}&&316cases&&Accuracy: 0.798\\
\multirow{4}{*}{\cite{193}}&\multirow{4}{*}{Hybrid}&Prostate Cancer&\multirow{4}{*}{H\&E}&Benign and malignant classification&Prostate core biopsy dataset&24859 cases&in house&AUC: 0.986\\
&&Skin Cancer&&Benign and malignant classification&Skin dataset&9,962 cases&in house&AUC: 0.986\\
&&Breast Cancer&&Detection of lymph node metastases&Breast dataset&9894 cases&\makecell{MSK breast cancer:\\http://thomasfuchslab.org/data/.}&AUC: 0.965\\
\cite{13}&Hybrid&Lung Cancer&H\&E&Diagnosis of lung cancer subtypes&Lung cancer dataset&939 cases&inhouse&Accuracy: 0.973\\
\cite{53}&Hybrid&Breast Cancer&IHC&\makecell{HER2 scoring (negative (0/1+),\\ equivocal (2+) and positive (3+))}&HER2 scoring dataset&1105 cases&inhouse&Accuracy: 0.8970\\
\multirow{2}{*}{\cite{177}}&\multirow{2}{*}{Hybrid}&\multirow{2}{*}{Colectoral Cancer}&\multirow{2}{*}{H\&E}&\multirow{2}{*}{\makecell{Prediction of normal\\and malignant tissues}}&CRC WSI Dataset I&173 cases&\multirow{2}{*}{inhouse}&Accuracy: 0.9231\\
&&&&&CRC WSI Dataset II&193 cases&&Accuracy: 0.9872\\
\multirow{2}{*}{\cite{54}}&\multirow{2}{*}{Hybrid}&\makecell{Gastrointestinal\\Celiac Disease}&\multirow{2}{*}{H\&E}&\makecell{Prediction of patients with\\celiac disease or being healthy}&Gastrointestinal dataset&413 cases&inhouse&Accuracy: 0.862\\
&&Breast Cancer&&Detection of lymph node metastases&Camelyon16 dataset&400 cases&\cite{211}&AUC: 0.9112\\
\multirow{3}{*}{\cite{55}}&\multirow{3}{*}{Hybrid}&Renal Cell Carcinoma&\multirow{3}{*}{H\&E}&\multirow{3}{*}{\makecell{subtyping and the detection\\of lymph node metastasis}}&RCC dataset&884 cases&https://portal.gdc.cancer.gov&AUC: 0.991\\
&&Non-small-cell Lung Cancer&&&NSCLC dataset&993 cases&https://cancerimagingarchive.net/datascope/cptac&AUC: 0.956\\
&&Breast Cancer&&&CAMELYON16 and CAMELYON17 dataset&899 cases&https://camelyon17.grand-challenge.org/Data&AUC: 0.936\\
\cite{12}&Hybrid&Prostate Cancer&H\&E&\makecell{Classifying cancer tissue\\into Gleason patterns}&\makecell{Prostate cANcer graDe Assessment\\(PANDA) challenge dataset}&11,000 cases&https://panda.grandchallenge.org/home/&Accuracy: 0.805\\
\multirow{2}{*}{\cite{10}}&\multirow{2}{*}{Clinical Studies}&\multirow{2}{*}{Breast Cancer}&\multirow{2}{*}{H\&E}&\multirow{2}{*}{\makecell{Determination of\\hormonal receptor status}}&\makecell{Australian Breast Cancer\\Tissue Bank (ABCTB) dataset}&2535 cases&https://abctb.org.au/abctbNew2/ACCESSPOLICY.pdf&AUC: 0.92\\
&&&&&\makecell{The Cancer Genome Atlas (TCGA) dataset}&1014 cases&https://portal.gdc.cancer.gov&AUC: 0.861\\
\cite{9}&Clinical Studies&Esophagus Cancer&H\&E&\makecell{Detection of cancerous and\\precancerous esophagus tissue}&Esophagus cancer dataset&180 cases&inhouse&Accuracy: 0.83\\
\cite{56}&Clinical Studies&Colorectal Cancer&H\&E&Prediction of colorectal cancer outcome&Colorectal cancer dataset&2473 cases&inhouse&\makecell{Ratio for poor versus\\good prognosis: 3.84}\\
\multirow{3}{*}{Kather \emph{et al.}~(\citeyear{57})}&\multirow{3}{*}{Clinical Studies}&\multirow{3}{*}{Gastrointestinal Cancer}&\multirow{3}{*}{H\&E}&\multirow{3}{*}{Prediction of microsatellite instability}&TCGA-STAD dataset&315 cases&\multirow{3}{*}{https://portal.gdc.cancer.gov/.}&AUC: 0.81\\
&&&&&TCGA-CRC-DX dataset&360 cases&&AUC: 0.84\\
&&&&&TCGA-CRC-KR dataset&378 cases&&AUC: 0.77\\
\multirow{3}{*}{\cite{58}}&\multirow{3}{*}{Clinical Studies}&\multirow{3}{*}{Lung Cancer}&\multirow{3}{*}{H\&E}&Classification of subtypes&\multirow{3}{*}{\makecell{The Cancer Genome Atlas (TCGA) dataset}}&\multirow{3}{*}{1634 cases}&\multirow{3}{*}{https://portal.gdc.cancer.gov/}&AUC: 0.97\\
&&&&\makecell{Prediction of mutation from\\ non-small cell lung cancer}&&&&\makecell{AUC of six of commonly mutated\\genes from 0.733 to 0.856}\\
\cite{44}&Clinical Studies&Breast Cancer&H\&E&Detection of lymph node metastases&CAMELYON16 dataset&400 cases&https://camelyon16.grand-challenge.org/&AUC: 0.994\\
\cite{61}&Clinical Studies&Prostate Cancer&H\&E&Prediction lymph node metastasis&Prostate cancer dataset&218 cases&inhouse&AUC: 0.68\\
\multirow{2}{*}{\cite{194}}&\multirow{2}{*}{Clinical Studies}&\multirow{2}{*}{Thyroid Cancer}&\multirow{2}{*}{H\&E}&\multirow{2}{*}{Prediction of BRAF mutation}&\makecell{ISBI 2017 Thyroid Tissue\\Microarray (TH-TMA17) dataset}&85 cases&Wang \emph{et al.}~(\citeyear{214})&AUC: 0.96\\
&&&&&TCGA-THCA dataset&444 cases&https://portal.gdc.cancer.gov/&AUC: 0.98\\
\multirow{2}{*}{\cite{195}}&\multirow{2}{*}{Clinical Studies}&\multirow{2}{*}{Breast Cancer}&\multirow{2}{*}{H\&E}&\multirow{2}{*}{\makecell{Prediction of HER2-positive breast\\cancer recurrence and metastasis risk}}&HER2-positive breast cancer dataset&127 cases&\multirow{2}{*}{https://github.com/bensteven2/HE\_breast\_recurrence}&AUC: 0.76\\
&&&&&The Cancer Genome Atlas (TCGA) dataset&123 cases&&AUC: 0.72\\
Li \emph{et al.}~(\citeyear{196})&Clinical Studies&Breast Cancer&H\&E&\makecell{Predicting biomarker of\\ pathological complete response\\ to neoadjuvant chemotherapy}&Breast cancer dataset&540 cases&inhouse&AUC: 0.847\\
\multirow{2}{*}{\cite{197}}&\multirow{2}{*}{Clinical Studies}&\multirow{2}{*}{Hepatocellular Carcinoma}&\multirow{2}{*}{H\&E}&\multirow{2}{*}{\makecell{Predicting survival after \\hepatocellular carcinoma resection}}&Discovery set&194 cases&inhouse&C-Indices: 0.78\\
&&&&&The Cancer Genome Atlas (TCGA) dataset&328 cases&https://portal.gdc.cancer.gov/&C-Indices: 0.70\\
\cite{198}&Clinical Studies&Bladder Cancer&H\&E&Identifying FGFR-activating mutations&\makecell{The Cancer Genome\\Atlas (TCGA) dataset}&418 cases&https://portal.gdc.cancer.gov/&AUC = 0.76\\
\multirow{2}{*}{\cite{199}}&\multirow{2}{*}{Clinical Studies}&\multirow{2}{*}{Bladder Cancer}&\multirow{2}{*}{H\&E}&\multirow{2}{*}{Prediction of molecular subtypes}&\makecell{The Cancer Genome Atlas (TCGA)\\Urothelial Bladder Carcinoma Dataset}&407 cases&https://portal.gdc.cancer.gov/&AUC = 0.89\\
&&&&&CCC-EMN cohort&16 cases&inhouse&AUC = 0.85\\
\bottomrule
\end{tabular}
\label{table1}
}%
\end{varwidth}}
\end{table}%
\normalsize

\subsection{Semi-Supervised Learning Paradigm}
\label{sec32}
Semi-supervised learning is a branch of machine learning that combines both supervised and unsupervised learning tasks and improves model performance by exploiting the information associated between tasks (Zhu \emph{et al.}~\citeyear{63}, Van \emph{et al.}~\citeyear{64}). In semi-supervised learning, only a small amount of labeled data is generally available, and besides that, a large amount of unlabeled data can be utilized for network training. Consequently, the main goal of semi-supervised learning is how to use these unlabeled data to improve the performance of the model trained with limited labeled data. Scenarios of the semi-supervised learning paradigm are very common in the field of pathological image analysis, both in diagnostic tasks and in segmentation tasks. Due to the expensive and time-consuming fine-grained annotation, pathologists often can only provide a small number of precise annotations for supervised training of the models, while a large amount of unannotated data cannot be used. Training deep models with only these limited labeled data can easily lead to over-fitting, thus significantly harming the performance and generalization of the models. In the semi-supervised learning paradigm, a large number of unlabeled images can be used to assist in training and thus further improve the performance, generalization, and robustness of the models.

In the past two decades, numerous semi-supervised learning algorithms have been proposed and widely used in the fields of natural image processing and pathological image analysis. The representative approaches in the field of semi-supervised learning are divided into five categories, namely pseudo-labelling-based approach (Section \ref{sec321}), consistency-based approach (Section \ref{sec322}), graph-based approach (Section \ref{sec323}), unsupervised-preprocessing approach (Section \ref{sec324}), and other approaches (Section \ref{sec325}). We introduce these methods below, respectively. For each category, we first describe their fundamental principles and then elaborate on their representative studies in the field of pathological image analysis. For a systematic review of the assumptions, concepts and representative methods of semi-supervised learning in the field of natural images, we recommend the review by Van \emph{et al.}~\citeyear{64}. Table \ref{table2} summarizes the detailed list of literatures in this section.

\subsubsection{Pseudo-labelling-based Approach}
\label{sec321}
\paragraph{Fundamental Principles}
The pseudo-labeling-based approach is a classical and well-known semi-supervised method (Zhu \emph{et al.}~\citeyear{63}), which mainly consists of two alternating processes, training and pseudo-labeling. Taking the classification problem as an example, in the training process, one or more classifiers are first trained in a supervised manner on the labeled data. The labeled data may be derived from the initial accurately labeled data or from the pseudo-labeled data from the previous iterations. In the pseudo-labeling process, all the unlabeled data are first predicted using the classifier trained in the previous process, and then the most confidently predicted portion of the data are selected for pseudo-labeling. Finally, these pseudo-labeled data are added to the labeled data for the next iteration. This process is repeated until no data with high confidence are found or all data are labeled.

The pseudo-labeling-based methods are firstly applied to the field of natural image processing and typically contain self-training methods (Lee \emph{et al.}~\citeyear{65}) and co-training methods (Blum \emph{et al.}~\citeyear{66}, Zhou \emph{et al.}~\citeyear{67}).

\paragraph{Study in Pathological Image Analysis}
In pathological image analysis, Singh \emph{et al.}~\citeyear{70} proposed a semi-supervised method of learning distance metrics from labeled data and performing label propagation for identifying the subtypes of nuclei, which was locally adaptive and could fully consider the heterogeneity of the data. Bulten \emph{et al.}~\citeyear{71} developed a deep learning system for Gleason scoring of prostate biopsies based on semi-supervised learning. They first trained the network on a small training dataset with pure Gleason scores, and then applied the trained network to other internal training datasets to set reference standards. Then, the labels were corrected and relabeled using reports from pathologists. Tolkach \emph{et al.}~\citeyear{72} used a pseudo-labeling-based semi-supervised strategy to train the CNN network to accomplish Gleason pattern classification. Jasiwal \emph{et al.}~\citeyear{87} proposed a semi-supervised method based on pseudo-labeling and entropy regularization for breast cancer pathological image classification. Shaw \emph{et al.}~\citeyear{73} extended the study of Yalniz \emph{et al.}~\citeyear{88} by proposing a semi-supervised teacher-student distillation method for the classification of colorectal cancer pathological images. Marini \emph{et al.}~\citeyear{69} proposed a deep pseudo-labeling-based semi-supervised learning method for strongly heterogeneous pathology data containing only a small number of local annotations. Their method consists of a high-volume teacher model and a small-volume student model, where the teacher model is automatically labeled with pseudo labels for the training of the student model. Cheng \emph{et al.}~\citeyear{74} proposed a semi-supervised learning framework based on a teacher-student model with similarity learning for the segmentation of breast cancer lesions containing a small number of annotations and noisy annotations.

\subsubsection{Consistency-based Approach}
\label{sec322}
\paragraph{Fundamental Principles}
The consistency-based semi-supervised learning approach is mainly based on the smoothing assumption. 
In the smoothing assumption, the prediction model should be robust to local perturbations within its input. 
This means that when we perturb the data points with a small amount of noise, the network's predictions for the perturbed data points and the clean original data points should be similar. 
In the implementation of deep neural networks, the consistency-based approach can be easily extended to a semi-supervised learning setup by directly adding unsupervised consistency loss functions to the original supervised loss functions. 
In the field of natural image processing, typical methods include $\pi$-model (Laine \emph{et al.}~\citeyear{75}), Temporal Ensembling model (Laine \emph{et al.}~\citeyear{75}), Mean Teachers (Tarvainen \emph{et al.}~\citeyear{76}) and UDA (Xie \emph{et al.}~\citeyear{77}).

\paragraph{Study in Pathological Image Analysis}
In pathological image analysis, Zhou \emph{et al.}~\citeyear{79} proposed a new Mean-teacher (MT) framework based on template-guided perturbation-sensitive sample mining. This framework consists of a teacher network and a student network. The teacher network is an integrated prediction network from K-times randomly augmented data, which is used to guide the student network to remain invariant to small perturbations at both feature and semantic levels. Su \emph{et al.}~\citeyear{78} proposed a novel global and local consistency loss and performed the nuclei classification task based on the Mean-Teacher framework.

\subsubsection{Graph-based Approach}
\label{sec323}
\paragraph{Fundamental Principles}
Methods of graph-based semi-supervised learning typically construct graphs to preserve the relationships of neighboring nodes, and use the graph transformations to simultaneously exploit information from labeled data and explore the underlying structure of unlabeled data. The key step of the graph-based semi-supervised learning methods is to construct a better graph to represent the original data structure. They usually define a graph on all data points (both labeled and unlabeled data points) and use weights to encode the similarity between pairs of the data points. In this way, the labeled information can be propagated through the graph to the unlabeled data points. For labeled data points, the predicted labels should match the true labels; similar data points defined by a similarity graph should have the same predictions. Graph-based semi-supervised methods are a relatively complex and long-developed field, and we recommend (Van \emph{et al.}~\citeyear{64}, Chong \emph{et al.}~\citeyear{82}) for a more thorough understanding.

\paragraph{Study in Pathological Image Analysis}
In pathological image analysis, Xu \emph{et al.}~\citeyear{80} proposed a new framework that combines a CNN with a semi-supervised regularization term. They first generated a hypothetical label for each unlabeled sample, then  proposed a graph-based smoothing term for regularization. Su \emph{et al.}~\citeyear{89} proposed an active learning and graph-based semi-supervised learning method for interactive cell segmentation. Inspired by the Temporal Ensembling model (Laine \emph{et al.}~\citeyear{75}), Shi \emph{et al.}~\citeyear{90} proposed a graph-based temporal ensembling model GTE. This method creates ensemble targets for both features and label predictions for each training sample, and encourages the model to form consistent predictions under different perturbations to exploit the semantic information of unlabeled data and improve the robustness of the model to noisy labels.

\subsubsection{Unsupervised-preprocessing-based Approach}
\label{sec324}
\paragraph{Fundamental Principles}
Unlike the previous approaches, unsupervised preprocessing-based approaches are typically dedicated to the unsupervised feature extraction, clustering (cluster-then-label), or initialization of the parameters of the subsequent supervised learning process (pre-training) from a large amount of unlabeled data. The most popular methods include autoencoders and their variants (Vincent \emph{et al.}~\citeyear{91,92}). Clustering is another method that enables adequate learning of the overall data distribution, thus many semi-supervised learning algorithms (Goldberg \emph{et al.}~\citeyear{93}, Demiriz \emph{et al.}~\citeyear{94}, Dara \emph{et al.}~\citeyear{95}) guide the subsequent classification process through clustering. The idea of the pre-training is to first pre-train a model using unsupervised methods with unlabeled data, and then use the parameters of this model as the initial parameters of the subsequent supervised training model, i.e., the subsequent supervised training is fine-tuned on the basis of these initial parameters. On this basis, the large number of unlabeled data can fully guide the subsequent classification models with limited labeled data thus improving the performance of semi-supervised learning (Erhan \emph{et al.}~\citeyear{96}).

\paragraph{Study in Pathological Image Analysis}
In pathological image analysis, Peikari \emph{et al.}~\citeyear{97} proposed a cluster-then-label semi-supervised learning method for identifying high-density regions in the data space and then utilized these regions to help support vector machines find decision boundaries. Lu \emph{et al.}~\citeyear{98} proposed a semi-supervised method based on feature extraction and pre-training for the WSI-level breast cancer classification task, which is the first work that relies on self-supervised feature learning using contrastive predictive coding for weakly supervised histopathological image classification. Koohbanani \emph{et al.}~\citeyear{125} proposed a joint framework of self-supervised learning and semi-supervised learning for pathological images. They proposed three pathology-specific self-supervised tasks, magnification prediction, magnification jigsaw prediction and hematoxylin channel prediction, to learn high-level semantic information and domain invariant information in pathological images. Srinidhi \emph{et al.}~\citeyear{124} also proposed a framework that combines self-supervised learning with semi-supervised learning. They first proposed the resolution sequence prediction (RSP) self-supervised auxiliary task to pre-train the model through unlabeled data, and then they performed fine-tuning of the model on the labeled data. After that they used the trained model from the above two steps as the initial weights of the model for further semi-supervised training based on the teacher-student consistency framework.

\subsubsection{Other Approaches}
\label{sec325}
\
\newline
Among semi-supervised learning, there are many other approaches, such as the methods based on generative adversarial networks (GAN) (Goodfellow \emph{et al.}~\citeyear{99,100}, Salimans \emph{et al.}~\citeyear{101}, Odena \emph{et al.}~\citeyear{102}, Dai \emph{et al.}~\citeyear{103}), Manifold-based methods (Belkin \emph{et al.}~\citeyear{104,105}, Weston \emph{et al.}~\citeyear{106}, Rifai \emph{et al.}~\citeyear{107,92}) and Association learning based methods (Haeusser \emph{et al.}~\citeyear{109}).

In pathological image analysis, Kapil \emph{et al.}~\citeyear{111} first used auxiliary classifier generative adversarial networks (AC-GAN) for the pathological image semi-supervised classification task and achieved favorable results. Cong \emph{et al.}~\citeyear{114} proposed to use a GAN-based semi-supervised learning method to accomplish the stain normalization problem for pathological images. Sparks \emph{et al.}~\citeyear{115} proposed a semi-supervised method based on epidemic learning to accomplish a content-based histopathological image retrieval task. Li \emph{et al.}~\citeyear{116} developed an Expectation-Maximization (EM)-based semi-supervised method for the semantic segmentation task of radical prostatectomy histopathological images. Su \emph{et al.}~\citeyear{110} proposed a new semi-supervised method based on association learning for pathological image classification task inspired by Haeusser \emph{et al.}~\citeyear{109}. Some studies (Foucart \emph{et al.}~\citeyear{117}) have also attempted to analyze the weaknesses and effectiveness of semi-supervised, noisy learning and weak label learning based on deep learning for pathological image analysis.

\begin{table}[!htbp]
\centering
\rotatebox[origin=c]{90}{%
\begin{varwidth}{\textheight}
\caption{List of literatures in the semi-supervised learning section.}
\footnotesize
\tabcolsep=6pt
\resizebox{.9\textwidth}{!}{%
\begin{tabular}{ccccccccc}
\toprule
\textbf{Reference}&\textbf{Approach}&\textbf{Disease Type}&\textbf{Staining}&\textbf{Task}&\textbf{Dataset}&\textbf{Dataset Scale}&\textbf{Dataset Link}&\textbf{Performance}\\
\midrule
\cite{70}&\makecell{Pseudo-\\labelling-based}&Breast Cancer&\makecell{3D fluorescence\\ microscopy}&\makecell{Identifying nuclear\\ phenotypes}&\makecell{Nuclei image\\ dataset}&984 images&Inhouse&Mean Accuracy: 0.8\\
\cite{71}&\makecell{Pseudo-\\labelling-based}&Prostate Cancer&H\&E&Gleason grading&Inhouse dataset&\makecell{5759 biopsies from\\ 1243 patients}&Inhouse&AUC = 0.99\\
\multirow{3}{*}{\cite{72}}&\multirow{3}{*}{\makecell{Pseudo-\\labelling-based}}&\multirow{3}{*}{Prostate Cancer}&\multirow{3}{*}{H\&E}&\makecell{Detection of prostate\\ cancer tissue}&\multirow{3}{*}{\makecell{The Cancer Genome\\Atlas Program\\ (TCGA) dataset}}&\multirow{4}{*}{1.67 million patches}&http://portal.gdc.cancer.gov&Accuracy = 0.967\\
&&&&\makecell{Gleason grading of \\prostatic adenocarcinomas}&&&https://zenodo.org/deposit/3825933&Accuracy = 0.98\\

\cite{87}&\makecell{Pseudo-\\labelling-based}&Breast Cancer&H\&E&\makecell{Detection of lymph\\ node metastases}&PatchCamelyon dataset& 327680 patches&https://camelyon16.grand-challenge.org/Data/&AUC = 0.9816\\
\cite{73}&\makecell{Pseudo-\\labelling-based}&Colorectal Cancer&H\&E&\makecell{Classification of 9 categories\\ of pathology patterns}&Public dataset& 100000 patches&\makecell{https://zenodo.org/record/1214456\#.YvyiX3ZByw4}&Mean Accuracy = 0.943\\

\multirow{2}{*}{\cite{69}}&\multirow{2}{*}{\makecell{Pseudo-\\labelling-based}}&\multirow{2}{*}{Prostate Cancer}&\multirow{2}{*}{H\&E}&\multirow{2}{*}{Gleason grading}&\makecell{Tissue MicroArray\\ dataset Zurich dataset}&886 cases&Inhouse&{$\kappa$-score: 0.7645}\\
&&&&&TCGA-PRAD dataset&449 cases&http://portal.gdc.cancer.gov&{$\kappa$-score: 0.4529}\\

\multirow{2}{*}{\cite{74}}&\multirow{2}{*}{\makecell{Pseudo-\\labelling-based}}&Breast Cancer&\multirow{2}{*}{H\&E}&\multirow{2}{*}{\makecell{Automated segmentation\\ of cancerous regions}}&CAMELYON16 dataset& 400 cases&https://camelyon16.grand-challenge.org/Data/&Dice: 93.76\\
&&Prostate Cancer&&&TVGH TURP dataset&71 cases&Inhouse&Dice: 77.24\\

\cite{79}&Consistency-based&-&\makecell{Liquid-based pap\\ test specimen}&\makecell{Cervical cell\\ instance segmentation}&\makecell{liquid-based Pap\\ test specimen dataset}&4439 cytoplasm&Inhouse&AJI: 73.45, MAP: 46.01\\
\multirow{2}{*}{\cite{78}}&\multirow{2}{*}{Consistency-based}&\multirow{2}{*}{-}&\multirow{2}{*}{H\&E}&\multirow{2}{*}{Nuclei classification}&MoNuseg dataset &22462 nuclei&\cite{215}&F1 score: 75.02 (5\% labels)\\
&&&&&Ki-67 nucleus dataset& 17516 nuclei&Inhouse&F1 score: 79.32 (5\% labels)\\

\multirow{3}{*}{\cite{124}}&\multirow{3}{*}{Consistency-based}&\multirow{3}{*}{\makecell{Breast Cancer,\\ Colorectal Cancer}}&\multirow{3}{*}{H\&E}&\makecell{Detection of\\ tumor metastasis}&BreastPathQ dataset&2579 patches&\cite{217}&TC: 0.876 (10\% labels)\\
&&&&Classification of tissue types&Camelyon16 dataset&399 WSIs&https://camelyon16.grand-challenge.org/Data/&AUC: 0.855 (10\% labels)\\
&&&&Quantification of tumor cellularity&Kather multiclass dataset&100K patches&Kather \emph{et al.}~(\citeyear{218})& Accuracy: 0.982 (10\% labels)\\

\cite{80}&Graph-based&-&Microscopy images&Neuron segmentation&\makecell{Neural morphology\\image dataset}&\makecell{2000 neuron regions\\with with annotations}&Inhouse&F1 score: 0.96 (40\% labels)\\
\cite{89}&Graph-based&-&Microscopy images&Cell segmentation&\makecell{Phase contrast microscopy\\ image dataset}&\makecell{Multiple sequences of\\ total 1404 frames}&http://www.celltracking.ri.cmu.edu/downloads.html.&TC: 0.9813\\

\multirow{2}{*}{\cite{90}}&\multirow{2}{*}{Graph-based}&Lung Cancer&\multirow{2}{*}{H\&E}&\multirow{2}{*}{Predictions of subtypes}&\multirow{2}{*}{\makecell{The Cancer Genome\\ Altas (TCGA)}}&2904 patches&\multirow{2}{*}{\makecell{http://portal.gdc.cancer.gov}}&Accuracy: 0.905 (20\% labels)\\
&&Breast Cancer&&&&1763 patches&&Accuracy: 0.895 (20\% labels)\\

\multirow{4}{*}{\cite{97}}&\multirow{4}{*}{\makecell{Unsupervised-\\preprocessing-based}}&\multirow{4}{*}{Breast Cancer}&\multirow{4}{*}{H\&E}&\multirow{4}{*}{\makecell{Identifying different\\breast tissue regions}}&\makecell{Pathology triaging\\image dataset}&4402 patches&\multirow{4}{*}{Inhouse}&{AUC: 0.86}\\
&&&&&\makecell{Nuclei figure\\ classification dataset}&30,000 figures&&AUC: 0.95\\

\cite{98}&\makecell{Unsupervised-\\preprocessing-based}&Breast Cancer&H\&E&\makecell{Benign and malignant\\ classification}&BACH dataset&400 cases&\makecell{BACH: Grand challenge on\\Breast Cancer histology images}&Accuracy: 0.95\\
\multirow{3}{*}{\cite{125}}&\multirow{3}{*}{\makecell{Unsupervised-\\preprocessing-based}}&Breast Cancer&\multirow{3}{*}{H\&E}&\makecell{Detection of\\tumor regions}&Camelyon16 dataset&399 slides&https://camelyon16.grand-challenge.org/Data/&{AUC: 0.817 (1\% labeled)}\\
&&Oral Squamous Cell Carcinoma&&\makecell{Prediction of metastases \\in the cervical lymph nodes}&LNM-OSCC dataset&217 slides&Inhouse&AUC: 0.806 (1\% labeled)\\
&&Colorectal Cancer&&Classification of tissue types&Kather multiclass dataset&100K patches&Kather \emph{et al.}~(\citeyear{218})&AUC: 0.903 (1\%labeled)\\

\multirow{3}{*}{\cite{124}}&\multirow{3}{*}{\makecell{Unsupervised-\\preprocessing-based}}&\multirow{3}{*}{\makecell{Breast Cancer,\\Colorectal Cancer}}&\multirow{3}{*}{H\&E}&\makecell{Detection of\\tumor metastasis}&BreastPathQ dataset &2579 patches&\cite{217}&TC: 0.876 (10\% labels)\\
&&&&Classification of tissue type&Camelyon16 dataset&399 WSIs&https://camelyon16.grand-challenge.org/Data/&AUC: 0.855 (10\% labels)\\
&&&&Quantification of tumor cellularity&Kather multiclass dataset &100K patches&Kather \emph{et al.}~(\citeyear{218})&ACC: 0.982 (10\% labels)\\

\cite{111}&GAN-based&Lung Cancer&\makecell{Ventana PD-L1\\(SP263) assay}&\makecell{Automated tumor\\proportion scoring}&\makecell{NSCLC needle\\biopsy dataset}&270 slides&Inhouse&\makecell{Ratio of the number\\of tumor positive\\ cell pixels to\\the total number of\\ tumor cell pixels: 0.94}\\
\multirow{2}{*}{\cite{114}}&\multirow{2}{*}{GAN-based}&Brain Cancer&\multirow{2}{*}{H\&E}&\multirow{2}{*}{Stain normalisation}&TCGA1 glioma cohort &22,229 images&\cite{219}&F1 score: 0.937\\
&&Breast Cancer&&&BreakHis database&7,909 images&\cite{220}&F1 score: 0.980\\

\cite{115}&\makecell{Manifold-\\learning-based}&Prostate Cancer&H\&E&Image retrieval&\makecell{Prostate\\histpathology dataset}&58 patients&Inhouse&SI: 0.14\\
\cite{116}&\makecell{Expectation-\\Maximization-based}&Prostate Cancer&H\&E&Semantic segmentation&Prostate dataset&\makecell{135 fully annotated and\\1800 weakly annotated tiles}&\cite{221}&AJI: 0.495\\
\multirow{2}{*}{\cite{110}}&\multirow{2}{*}{\makecell{Association-\\learning-based}}&\multirow{2}{*}{Breast Cancer}&\multirow{2}{*}{H\&E}&\multirow{2}{*}{\makecell{Classification of cancerous\\and non-cancerous slides}}&\makecell{Bioimaging 2015\\challenge dataset}&285 images&\cite{222}&F1 score: 0.75\\
&&&&&BACH dataset&400 images&\cite{209}&F1 score: 0.77\\

\bottomrule
\end{tabular}
\label{table2}
}%
\end{varwidth}}
\end{table}%
\normalsize
  
\subsection{Self-Supervised Learning Paradigm}
\label{sec33}
Unlike the former two paradigms, the self-supervised learning paradigm does not perform the classification or segmentation of pathological images directly, but in a two-stage 'pre-training and fine-tuning' approach. Due to the small number of annotated pathological images, it is not enough to use these data to directly train the model. Therefore, the self-supervised learning paradigm aims to first learn effective feature representations from a large amount of unlabeled data, which is called the pre-training process. Afterwards, the feature representations learned in the self-supervised auxiliary tasks are used to be transferred to train the downstream tasks using limited labeled data, which is called the fine-tuning process. In this way, good feature representations can effectively help the model to achieve good results even if it is trained with only a small amount of labeled data.

The process of pre-training, i.e., the learning process of good feature representations, is the key to self-supervised learning. Typically, self-supervised learning learns good feature representations by performing self-supervised auxiliary tasks. In a self-supervised auxiliary task, certain inherent properties of the unlabeled data are first used to generate supervised signals, and then the network is trained by these self-supervised signals. Different studies usually focus on designing different self-supervised auxiliary tasks to perform feature representation learning efficiently. According to the properties of the auxiliary tasks, existing self-supervised learning paradigms can be mainly classified into predictive self-supervised learning, generative self-supervised learning, and contrastive self-supervised learning. Predictive self-supervised learning learns good feature representations by constructing the auxiliary tasks as classification problems with unlabeled data; generative self-supervised learning learns good feature representations by reconstructing the input images; and contrastive self-supervised learning learns good feature representations by learning to distinguish between similar samples (positive samples) and dissimilar samples (negative samples). For a systematic review of self-supervised methods in the natural image domain and medical image domain, we recommend the reviews by Liu \emph{et al.}~\citeyear{118} and Shurrab \emph{et al.}~\citeyear{119}.

In this section, we provide a detailed review of the studies on self-supervised learning for pathological image analysis. Currently, some studies focus on proposing innovative self-supervised frameworks for pathological images (we call them study on novel self-supervised frameworks), while others attempt to apply existing self-supervised learning methods to pathological image analysis (we call them study on application of self-supervised frameworks). We introduce studies on novel self-supervised frameworks in Section \ref{sec331}, where we focus on predictive self-supervised learning, generative self-supervised learning, contrastive self-supervised learning, and hybrid self-supervised learning and their state-of-the-art approaches in pathological image analysis. We introduce the study on application of self-supervised frameworks in Section \ref{sec332}. Table \ref{table3} summarizes a detailed list of literatures in this section.

\subsubsection{Study on Novel Self-supervised Frameworks}
\label{sec331}
\paragraph{Predictive Self-supervised Learning Approach}
\label{sec3311}
\subparagraph{Fundamental Principles}
Predictive self-supervised learning learns good feature representations by constructing the auxiliary tasks as classification problems with unlabeled data, and the class labels for classification are constructed from the unlabeled data itself. Currently, predictive self-supervised auxiliary tasks widely applied in natural image processing are relative position prediction (Doersch \emph{et al.}~\citeyear{120}), solving Jigsaw puzzles (Noroozi \emph{et al.}~\citeyear{121}), and rotation angle prediction (Gidaris \emph{et al.}~\citeyear{122}), etc.

\subparagraph{Study in Pathological Image Analysis}
In the field of pathological image processing, Sahasrabudhe \emph{et al.}~\citeyear{123} proposed the auxiliary task of predicting patch magnification for cell nuclei segmentation. Their main idea is that given WSIs of different magnification classes (e.g., 5$\times$, 10$\times$, 20$\times$), they first obtained patches of different magnifications from them and then predicted the magnification class of those patches by examining the size and texture of the cell nuclei in the patches. Srinidhi \emph{et al.}~\citeyear{124} proposed the resolution sequence prediction (RSP) auxiliary task. First they used patches with different magnifications to construct different combinations of resolution sequences, and then trained the network to predict the order of the resolution sequences. Koohbanani \emph{et al.}~\citeyear{125} proposed magnification prediction and solving magnification puzzles auxiliary tasks for pathological images. They first trained the network to accurately predict the magnification category, and then trained the network to predict the order of the patches with different magnifications.

\paragraph{Generative Self-supervised Learning Approach}
\label{sec3312}
\subparagraph{Fundamental Principles}
Generative self-supervised learning learns good feature representations by reconstructing the input images. They argue that the image itself is a useful self-supervised information and that the network can learn the potential feature representations of the generated image during the image reconstruction process. In natural image processing, autoencoders (Goodfellow \emph{et al.}~\citeyear{126}) are representative of early work on generative self-supervised feature representation learning. 
Later, denoising autoencoders (Vincent \emph{et al.}~\citeyear{91}) enhanced the feature representation capability of the model by introducing noise. Subsequently, researchers proposed a series of reconstructive self-supervised auxiliary tasks, including inpainting (Pathak \emph{et al.}~\citeyear{128}), colorization (Zhang \emph{et al.}~\citeyear{129}), patch shuffling and restoration (Chen \emph{et al.}~\citeyear{130}, Zhou \emph{et al.}~\citeyear{131}) to further enhance the feature representation capability of the network and achieved promising results. On the other hand, a series of GAN-based models (e.g., DCGAN \citeyear{132}, BiGAN \citeyear{133}) have also been used to perform self-supervised representation learning. In the latest self-supervised studies on natural images, a series (e.g., BEiT \citeyear{134}, MAE \citeyear{135}, PeCo \citeyear{136}, etc.) of self-supervised studies based on masked image blocks and reconstruction using Transformer achieved the highest performance, which is expected to start a new wave of research on reconstruction-based self-supervised representation learning.

\subparagraph{Study in Pathological Image Analysis}
In pathological image analysis, Muhammad \emph{et al.}~\citeyear{137} proposed a new deep convolutional autoencoder-based clustering model to learn the feature representations of pathological images. 
Mahapatra \emph{et al.}~\citeyear{138} incorporated semantic information into a GAN-based generative model for self-supervised feature representation learning and used it for the stain normalization task of pathological images. 
Quiros \emph{et al.}~\citeyear{154,155} designed GANs for pathological images to extract key feature representations of tissues. 
Boyd et al \emph{et al.}~\citeyear{139} proposed a new generative auxiliary task which performs representation learning by extending the view of image patches. 
Hou et al \emph{et al.}~\citeyear{200} proposed a sparse convolutional autoencoder (CAE) for simultaneous nuclei detection and feature extraction in histopathological images. 
Koohbanani \emph{et al.}~\citeyear{125} proposed the hematoxylin channel prediction auxiliary task, where they used hematoxylin and eosin (H\&E) stained images to predict the hematoxylin channel pixel by pixel. 

\paragraph{Contrastive Self-supervised Learning Approach}
\label{sec3313}
\subparagraph{Fundamental Principles}
The contrastive self-supervised approach is one of the most popular self-supervised paradigms, which focuses on learning good feature representations by encouraging the model to learn to distinguish between similar samples (positive samples) and dissimilar samples (negative samples).

Contrast predictive coding (CPC) (Van \emph{et al.}~\citeyear{141}) is an early contrastive self-supervised method applied to natural image processing whose goal is to maximize the mutual information between patches (positive samples) from the same image and minimize the mutual information between patches (negative samples) from different images within a mini-batch. Typical subsequent studies have been devoted to constructing negative samples. MoCo (He \emph{et al.}~\citeyear{143}) is a momentum-based contrastive self-supervised framework, which is mainly based on the ideas of dynamic dictionary-lookup and queues. 
SimCLR (Chen \emph{et al.}~\citeyear{142}) is a simple contrastive learning framework that aims to maximize the cosine similarity between two augmented views of the same image (positive samples) and minimize the similarity between different images in a minibatch (negative samples).

These methods rely heavily on a large number of negative samples since only positive samples will easily lead to model degeneration, i.e., mapping the features of all samples to an identical vector. However, recent studies have shown that negative samples are not necessary. Caron \emph{et al.}~\citeyear{144} introduced clustering into contrastive learning, thus eliminating the need for negative samples. Chen \emph{et al.}~\citeyear{152} explored stop-gradient operation applied to siamese networks without the need for a large number of negative samples. Grill \emph{et al.}~\citeyear{145}, Caron \emph{et al.}~\citeyear{146} proposed a self-supervised learning model based on a teacher-student knowledge distillation framework that achieves state-of-the-art performance without any negative samples.

\subparagraph{Study in Pathological Image Analysis}
In pathological image analysis, Xie \emph{et al.}~\citeyear{147} employed patches from different magnifications as positive samples and patches from different magnifications as negative samples and constructed scale-wise triplet loss to perform contrastive learning for the nuclei segmentation. 
Chhipa \emph{et al.}~\citeyear{148} proposed Magnification Prior Contrastive Similarity (MPCS) to construct contrastive loss. 
Xu \emph{et al.}~\citeyear{153} proposed a self-supervised Deformation Representation Learning (DRL) framework to learn semantic features from unlabeled pathological images. 
They used mutual information to train the network to distinguish original histopathological images from those deformed in local structure, while consistent global contextual information was maintained using noise contrastive estimation (NCE). 
Wang \emph{et al.}~\citeyear{149} proposed Transpath based on the BYOL framework \citeyear{145}. They first collected the current largest histopathological image dataset for self-supervised pre-training, which includes about 2.7 million images from 32529 WSIs. Then they proposed a hybrid framework combining CNN and Transformer to extract both local structural features and global contextual features, and proposed a TAE module to further enhance the feature extraction capability.

\paragraph{Hybrid Self-supervised Learning Approach}
\label{sec3314}
Many studies have also presented hybrid self-supervised methods for pathological images. Abbet \emph{et al.}~\citeyear{150} proposed a combination of generative and contrastive self-supervised representation learning method for pathological images. 
They first applied colorization as a generative auxiliary task. Then, they constructed the contrastive loss using spatially neighboring patches as positive samples and distant patches as negative samples. 
Yang \emph{et al.}~\citeyear{151} also proposed a self-supervised representation method combining generative and contrastive approaches for pathological images. 
They first proposed a generative-based self-supervised task called cross-stain prediction, in which they defined two encoders and decoders to predict the E-channel and H-channel, respectively, and then they used the encoders trained in the previous task to perform further contrastive training. 

\subsubsection{Study on Applications of Self-supervised Frameworks}
\label{sec332}
\
\newline
In addition to studies that aim to propose innovative self-supervised frameworks for pathological images, more studies have attempted to apply existing self-supervised learning methods to various pathological image analysis tasks. Chen \emph{et al.}~\citeyear{156} proposed an end-to-end multimodal fusion framework for histopathological images and genomic data for survival prognosis prediction, in which they used contrastive predictive coding (CPC) pre-trained self-supervised features for initialization of the network model. Ciga \emph{et al.}~\citeyear{157} showed through extensive experiments that using self-supervised pre-training methods can yield better features to improve performance on several downstream tasks. They found that the success of contrastive self-supervised pre-training methods depended heavily on the diversity of the unlabeled training set rather than the number of images. On the other hand, positive and negative samples that are visually significantly different facilitate contrastive self-supervised learning, while positive and negative sample that contain only minor differences but are generally similar (e.g., normal patches versus patches containing only a small percentage of tumor regions) are not conducive to contrastive learning. However, this is uncommon in natural images, so it is particularly important to design targeted self-supervised tasks for the characteristics of pathological images. Tellez \emph{et al.}~\citeyear{163} used the variational autoencoder \citeyear{165}, contrastive learning \citeyear{166} and BiGAN \citeyear{133} for the compression of gigapixel pathological images and evaluated the performance on a synthetic dataset and two public histopathology datasets, respectively, achieving promising results. Stacke \emph{et al.}~\citeyear{164} investigated how SimCLR \citeyear{142} could be extended for pathological images to learn useful feature representations. They systematically compared the differences between ImageNet data and histopathology data and how this affected the goals of self-supervised learning, and pointed out the impact that designing for different self-supervised goals would have on the results. Chen \emph{et al.}~\citeyear{167} comprehensively compared the performance of ImageNet pre-trained features, SimCLR pre-trained features, and DINO \citeyear{146} pre-trained features in weakly supervised classification and fully supervised classification tasks for histopathological images. They found that the DINO-based knowledge distillation framework could better learn effective and interpretable features in pathological images.

Saillard \emph{et al.}~\citeyear{158} and Dehaene \emph{et al.}~\citeyear{161} used the MoCo V2 \citeyear{159} self-supervised learning method to train pathological images and the experimental results showed that the results using the self-supervised pre-trained features were consistently better than those using features pre-trained on ImageNet under the same conditions. 
Lu \emph{et al.}~\citeyear{98}, Zhao \emph{et al.}~\citeyear{192}, and Li \emph{et al.}~\citeyear{191} used contrastive predictive coding (CPC) \citeyear{141}, VAE-GAN \citeyear{162}, and SimCLR \citeyear{142} self-supervised pre-trained features for weakly supervised WSI classification, respectively, and achieved the current state-of-the-art performance. 
Koohbanani \emph{et al.}~\citeyear{125} developed a semi-supervised learning framework facilitated by self-supervised learning with a multi-task learning approach for training, i.e., training with a small amount of labeled data as the main task and self-supervised tasks as auxiliary tasks. In their study, they also compared the effectiveness of various commonly used pathology-agnostic self-supervised auxiliary tasks (including rotation, flipping, auto-encoder, real/fake prediction, domain prediction, etc.) to facilitate semi-supervised learning. Srinidhi \emph{et al.}~\citeyear{124} also attempted to use self-supervised pre-trained features to enhance semi-supervised learning. They first proposed the resolution sequence prediction (RSP) self-supervised auxiliary task to pre-train the model through unlabeled data, and then they fine-tuned the model on the labeled data. After that, they used the trained model from the above two steps as the initial weights of the model for further semi-supervised training based on the teacher-student consistency framework.

In addition, self-supervised learning has been used for a variety of other pathological tasks, such as pathological image retrieval (Shi \emph{et al.}~\citeyear{169}, Yang \emph{et al.}~\citeyear{170}), active learning (Zheng \emph{et al.}~\citeyear{171}), and molecular signature prediction (Ding \emph{et al.}~\citeyear{172}, Fu \emph{et al.}~\citeyear{173}, Kather \emph{et al.}~\citeyear{174}), etc.

\begin{table}[!htbp]
  \centering
  \caption{List of literatures in the self-supervised learning section.}
  \footnotesize
  \tabcolsep=3pt
  \resizebox{\textwidth}{.45\textheight}{%
  \begin{tabular}{m{12em}<{\centering}m{6em}<{\centering}m{10em}<{\centering}m{4em}<{\centering}m{17em}<{\centering}m{7em}<{\centering}m{29em}<{\centering}m{13em}<{\centering}m{14em}<{\centering}m{15em}<{\centering}}
  \toprule
  \textbf{Reference}&\textbf{Approach}&\textbf{Disease Type}&\textbf{Staining}&\textbf{Dataset}&\textbf{Dataset Scale}&\textbf{Dataset Link}&\textbf{Self-supervised Method}&\textbf{Downstream Task}&{\textbf{Downstream Performance}}\\
  \midrule
  \cite{123}&Predictive&-&H\&E&MoNuSeg database&1,125,737 tiles&\cite{224}&Identification of the magnification levels for tiles&Nuclei segmentation&{AJI: 0.5354, AHD: 7.7502, Dice: 0.7477}\\\hline
  \multirow{4}{*}{\cite{124}}&\multirow{4}{*}{Predictive}&\multirow{4}{*}{\makecell{Breast Cancer,\\Colorectal Cancer}}&\multirow{4}{*}{H\&E}&BreastPathQ dataset&2579 patches&\cite{217}&\multirow{4}{*}{\makecell{Predicting the\\resolution sequences}}&\makecell{Detection of\\tumor metastasis}&TC: 0.876 (10\% labels)\\
  &&&&Camelyon16 dataset&399 WSIs&\url{https://camelyon16.grand-challenge.org/Data/}&&Classification of tissue types&{AUC: 0.855 (10\% labels)}\\
  &&&&Kather multiclass dataset&100K patches&Kather \emph{et al.}~(\citeyear{218})&&Quantification of tumor cellularity &{Accuracy: 0.982 (10\% labels)}\\\hline
  
  \multirow{4}{*}{\cite{125}}&\multirow{4}{*}{Predictive}&Breast Cancer&\multirow{4}{*}{H\&E}&Camelyon16 dataset&399 slides&\url{https://camelyon16.grand-challenge.org/Data/}&\multirow{4}{*}{\makecell{Magnification prediction and\\ solving magnification puzzles}}&Detection of tumor regions&AUC: 0.817 (1\% labeled)\\
  &&oral Squamous Cell Carcinoma&&LNM-OSCC dataset&217 slides&Inhouse&&\makecell{Prediction of metastases in\\ the cervical lymph nodes}&AUC: 0.806 (1\% labeled)\\
  &&Colorectal Cancer&&Kather multiclass dataset&100K patches&Kather \emph{et al.}~(\citeyear{218})&&Classification of tissue types&AUC: 0.903 (1\%labeled)\\
  \hline
  \cite{137}&Generative&Cholangi-ocarcinoma&H\&E&Intrahepatic cholangiocarcinoma (ICC) dataset&246 patients&Inhouse&Deep clustering convolutional autoencoder&Subtyping of cholangiocarcinoma&{CHI: 3863(5 clusters) and 4314 (clutsering weight = 0.2)}\\\hline
  \multirow{3}{*}{\cite{138}}&\multirow{3}{*}{Generative}&Breast Cancer&\multirow{3}{*}{H\&E}&CAMELYON16 dataset&100, 000 patches&\url{https://camelyon16.grand-challenge.org/Data/}&\multirow{3}{*}{\makecell{Using pre-trained networks\\ for semantic guidance}}&\multirow{3}{*}{Stain normalization}&\multirow{3}{*}{Average AUC: 0.9320}\\
  &&Breast Cancer&&CAMELYON17 dataset&100, 000 patches&\url{https://camelyon16.grand-challenge.org/Data/}, inhouse&&&\\
  \hline
  \multirow{4}{*}{\cite{154}}&\multirow{4}{*}{Generative}&\multirow{2}{*}{Colorectal Cancer}&\multirow{4}{*}{H\&E}&National Center for Tumor diseases (NCT) dataset&86 slides&\url{https://zenodo.org/record/1214456\#.Yvzd-nZBxhE}&\multirow{4}{*}{\makecell{Using Generative Adversarial\\ Networks (GANs) to capture\\ key tissue features and\\ structure information}}&\multirow{4}{*}{\makecell{Count of cancer,\\ lymphocytes, or stromal cells}}&\multirow{2}{*}{FID: 16.65}\\
  &&Breast Cancer&&Netherlands Cancer Institute (NKI) dataset and Vancouver General Hospital (VGH) dataset&576 tissue micro-arrays (TMAs)&\cite{225}&&&FID: 32.05 \\
  \hline
  
  \multirow{4}{*}{\cite{155}}&\multirow{4}{*}{Generative}&Breast Cancer&\multirow{4}{*}{H\&E}&Netherlands Cancer Institute (NKI, Netherlands) and Vancouver General Hospital (VGH, Canada) cohorts&Total of 576 patients&\cite{225}&\multirow{4}[1]{*}{\makecell{Presenting an adversarial\\learning model to extract\\ feature representations\\ of cancer tissue}}&\multirow{4}{*}{\makecell{Classifying tissue types and\\ predicting the presence of tumor\\ in Whole Slide Images (WSIs)\\ using multiple instance learning (MIL)}}&\multirow{4}{*}{\makecell{AUC: 0.97 and\\ Accuracy: 0.85; AUC: 0.98\\ and Accuracy: 0.94}}\\
  &&Colon cancer&&National Center for Tumor diseases (NCT, Germany) dataset&100K tissue tiles&\url{https://zenodo.org/record/1214456\#.Yvzd-nZBxhE}&&&\\
  &&Lung Cancer&&TCGA LUAD, LUSC dataset&1184 patients&\url{http://portal.gdc.cancer.gov}&&&\\
  \hline
  
  \multirow{3}{*}{\cite{139}}&\multirow{3}{*}{Generative}&Breast Cancer&\multirow{3}{*}{H\&E}&CAMELYON17 dataset&500 slides&\url{https://camelyon16.grand-challenge.org/Data/}&\multirow{3}{*}{Visual field expansion}&Binary classification of tiles into metastatic and non-metastatic classes&Accuracy: 0.8569\\
  &&Colorectal Cancer&&CRC benchmark dataset&100K image tiles&\url{https://doi.org/10.5281/zenodo.1214456}&&Classification of tiles into the 9 tissue types&{Accuracy: 0.8511}\\
  \hline
  
  \multirow{5}{*}{\cite{125}}&\multirow{5}{*}{Generative}&Breast Cancer&\multirow{5}{*}{H\&E}&Camelyon16 dataset&399 slides&\url{https://camelyon16.grand-challenge.org/Data/}&\multirow{5}{*}{\makecell{Hematoxylin channel\\prediction auxiliary task}}&Detection of tumor regions&AUC: 0.817 (1\% labeled)\\
  &&Oral Squamous Cell Carcinoma&&LNM-OSCC dataset&217 slides&Inhouse&&Prediction of metastases in the cervical lymph nodes&AUC: 0.806 (1\% labeled)\\
  &&Colorectal Cancer&&Kather multiclass dataset&100K patches&Kather \emph{et al.}~(\citeyear{218})&&Classification of tissue types&AUC: 0.903 (1\%labeled)\\
  \hline

  \multirow{9}{*}{\cite{200}}&\multirow{9}{*}{Generative}&\multirow{9}{*}{-}&\multirow{9}{*}{H\&E}&Self-collected lymphocyte classification dataset&1785 images&Inhouse&\multirow{9}{*}{\makecell{Sparse Convolutional\\Autoencoder (CAE)}}&\multirow{9}{*}{Nucleus detection}&Nucleus Classification: Lymphocyte Classification AUC 0.7856\\
  &&&&Nuclear shape and attribute classification dataset&2000 images&\cite{226}&&& Nuclear Attribute \&Shape AUC 0.8788\\
  &&&&CRCHistoPhenotypes nucleus detection dataset&100 images&\cite{215}&&&Nucleus detection: F-measure: 0.8345\\
  &&&&MICCAI 2015 nucleus segmentation challenge dataset&763 images&\makecell{https://wiki.cancerimagingarchive.net/\\pages/viewpage.action?pageId=20644646}&&&Lymphocyte classification: AUC 0.7856\\
  &&&&TCGA lung cancer dataset&0.5 million images&\url{https://cancergenome.nih.gov/}&&&{Nucleus segmentation: DICE: 0.8362}\\
  \hline
  \cite{147}&Contrastive&-&H\&E&MoNuSeg dataset&44 images&\cite{227}&Scale-wise triplet learning and count ranking&Nuclei segmentation&AJI: 0.7063\\
  \hline
  \cite{148}&Contrastive&Breast Cancer&H\&E&BreakHis dataset&7909 images&\cite{220}&Magnification prior contrastive similarity&Classifying histopathological images&{Mean Accuracy: 0.9233}\\
  \hline
  \multirow{3}{*}{\cite{153}}&\multirow{3}{*}{Contrastive}&Breast Cancer&\multirow{3}{*}{H\&E}&MICCAI 2015 Gland Segmentation Challenge (GLaS) dataset&165 images&\cite{228}&\multirow{3}{*}{\makecell{Deformation \\representation learning}}&Gland segmentation&\multirow{3}{*}{\makecell{F1-score 0.900, Accuracy\\ 0.8548 (10\% labeled)}}\\
  &&Colon Cancer&&Patch Camelyon (PCam) image classification dataset&327,680 patches&\cite{229}&&Semi-supervised classification&\\
  \hline
  \cite{149}&Contrastive&Liver, Renal, Colorectal, Prostatic, Pancreatic, and Cholangio Breast Cancers&H\&E&Multiple histopathological image datasets including MHIST, NCT-CRC-HE, PatchCamelyon dataset&2.7 million images&\url{https://github.com/Xiyue-Wang/TransPath}&Contrastive learning like BYOL (Bootstrap your own latent: a new approach to self-supervised learning)&Histopathological image classification tasks&{F1-score: 0.8993, 0.9582, 0.8983 on MHIST, NCT-CRC-HE, PatchCamelyon dataset}\\
  \hline
  \cite{150}&Generative + Contrastive&Colorectal Cancer&H\&E&Clinicopathological dataset&660 WSIs&Inhouse&Colorization, Image reconstrucation and Contrastive learning&Survival analysis&{C-Index: 0.6943}\\
  \hline
  \cite{151}&Generative + Contrastive&Colorectal Cancer&H\&E&NCTCRC-HE-100K dataset&100K images&\url{https://zenodo.org/record/1214456\#.Yvzd-nZBxhE}&Cross-stain prediction, Contrastive training&Nine-class classification of histopathological images&{Accuracy of eight-class classification with only 1,000 labeled data: 0.915}\\
  \hline
  \cite{156}&Application&Glioma and Cell Carcinoma&H\&E&The Cancer Genome Atlas (TCGA) dataset&1505 images&\url{http://portal.gdc.cancer.gov}&Contrastive predictive coding (CPC)&Survival prognosis prediction&{C-Index: 0.826}\\
  \hline
  \cite{157}&Application&Multiple Types&H\&E&Out of the total 57 datasets from various institutions&A large number of images&\url{https://github.com/ozanciga/self-supervised-histopathology}&Contrastive learning&Classification, Regression, and Segmentation&{Multiple results}\\
  \hline
  \multirow{3}{*}{\cite{163}}&\multirow{3}{*}{Application}&\multirow{3}{*}{Breast Cancer}&\multirow{3}{*}{H\&E}&Camelyon16 dataset&400 WSIs&\url{https://camelyon16.grand-challenge.org/Data/}&\multirow{3}{*}{\makecell{Variational autoencoder,\\ Contrastive learning and BiGAN}}&Predicting the presence of metastasis&AUC: 0.725\\
  &&&&TUPAC16 dataset&492 WSIs&\cite{230}&&\makecell{Predicting tumor\\ proliferation speed}&\makecell{Spearman correlation:\\ 0.522}\\
  \hline
  \multirow{4}{*}{\cite{164}}&\multirow{4}{*}{Application}&\multirow{4}{*}{Multiple Types}&\multirow{4}{*}{H\&E}&Camelyon16 dataset&400 slides&\multirow{4}{*}{\url{https://github.com/k-stacke/ssl-pathology}}&\multirow{4}{*}{Contrastive learning}&\multirow{4}{*}{Binary tumor classification}&\multirow{4}{*}{Multiple results}\\
  &&&&AIDA-LNSK dataset&96 slides&&&&\\
  &&&&Multidata (samples from 60 publicly available datasets)&A large number of images&&&&\\
  \hline
  \multirow{2}{*}{\cite{167}}&\multirow{2}{*}{Application}&Colorectal Cancer&\multirow{2}{*}{H\&E}&CRC-100K dataset&100K images&\cite{231}&\multirow{2}{*}{Contrastive learning}&Weakly-supervised cancer subtyping&AUC: 0.886\\
  &&Breast Cancer&&BreastPathQ dataset&2766 patches&\cite{232}&&Patch-level tissue phenotyping&{AUC: 0.987}\\
  \hline
  \multirow{2}{*}{\cite{158}}&\multirow{2}{*}{Application}&Colorectal Cancer&\multirow{2}{*}{H\&E}&TCGA-CRC dataset&555 patients&\multirow{2}{*}{\url{http://portal.gdc.cancer.gov}}&\multirow{2}{*}{Contrastive learning}&\multirow{2}{*}{Microsatellite instability}&AUC: 0.92\\
  &&Gastric Cancer&&TCGA-Gastric dataset&375 patients&&&&{AUC: 0.83}\\
  \hline
  \multirow{3}{*}{\cite{161}}&\multirow{3}{*}{Application}&Colorectal Cancer&\multirow{3}{*}{H\&E}&Camelyon16 dataset&400 slides&\url{https://camelyon16.grand-challenge.org/Data/}&\multirow{3}{*}{Contrastive learning}&Predicting lymph node metastasis in Breast Cancer&AUC: 0.987\\
  &&Breast Cancer&&TCGA-COAD dataset&461 slides&\cite{233}&&Colorectal Cancer subtyping&\makecell{AUC: 0.882 (CMS1) and\\ AUC: 0.829 (CMS3)}\\
  \hline
  \cite{98}&Application&Breast Cancer&H\&E&BACH dataset&400 cases&\cite{209}&Contrastive predictive coding (CPC)&classification and localization of clinically relevant histopathological classes&Accuracy: 0.95\\
  \hline
  \cite{192}&Application&Colon Adenocarcinoma&H\&E&The Cancer Genome Atlas (TCGA) dataset&425 patients&\url{http://portal.gdc.cancer.gov}&Variational Auto Encoder and Generative Adversial Network (VAE-GAN)&Predicting lymph node metastasis&{Accuracy: 0.6761}\\
  \hline
  \multirow{3}{*}{\cite{191}}&\multirow{3}{*}{Application}&Breast Cancer&\multirow{3}{*}{H\&E}&Camelyon16 dataset&400 cases&\url{https://camelyon16.grand-challenge.org/Data/}&\multirow{3}{*}{Contrastive learning}&Detection of lymph node metastases&Accuracy: 0.8992\\
  &&Lung Cancer&&TCGA lung cancer dataset&1054 cases&\url{https://www.cancer.gov/about-nci/organization/ccg/research/structural-genomics/tcga}&&Diagnosis of lung cancer subtypes&Accuracy: 0.9571\\
  \hline
  \multirow{4}{*}{\cite{125}}&\multirow{4}{*}{Application}&Breast Cancer&\multirow{4}{*}{H\&E}&Camelyon16 dataset&399 slides&\url{https://camelyon16.grand-challenge.org/Data/}&\multirow{4}{*}{\makecell{Magnification prediction,\\ JigMag prediction and\\ Hematoxylin channel prediction}}&Detection of tumor regions&AUC: 0.817 (1\% labeled)\\
  &&Oral Squamous Cell Carcinoma&&LNM-OSCC dataset&217 slides&Inhouse&&\makecell{Prediction of metastases in\\ the cervical lymph nodes}&AUC: 0.806 (1\% labeled)\\
  &&Colorectal Cancer&&Kather multiclass dataset&100K patches&Kather \emph{et al.}~(\citeyear{218})&&Classification of tissue types&AUC: 0.903 (1\%labeled)\\
  \hline
  \multirow{4}{*}{\cite{124}}&\multirow{4}{*}{Application}&\multirow{4}{*}{\makecell{Breast Cancer,\\ Colorectal Cancer}}&\multirow{4}{*}{H\&E}&BreastPathQ dataset&2579 patches&\cite{217}&\multirow{4}{*}{Resolution sequence prediction}&Detection of tumor metastasis&TC: 0.876 (10\% labels)\\
  &&&&Camelyon16 dataset&399 WSIs&\url{https://camelyon16.grand-challenge.org/Data/}&&Classification of tissue types&{AUC: 0.855 (10\% labels)}\\
  &&&&Kather multiclass dataset&100K patches&Kather \emph{et al.}~(\citeyear{218})&&Quantification of tumor cellularity &ACC: 0.982 (10\% labels)\\
  \hline
  \multirow{2}{*}{\cite{171}}&\multirow{2}{*}{Application}&\multirow{2}{*}{Colon Cancer }&\multirow{2}{*}{H\&E}&MICCAI 2015 Gland Segmentation Challenge (GlaS) dataset&165 images&\cite{228}&\multirow{2}{*}{\makecell{Variational Auto\\Encoder (VAE)}}&\multirow{2}{*}{\makecell{Active learning in\\ biomedical image segmentation}}&\multirow{2}{*}{\makecell{F1 score: 0.909,\\0.9252 (30\% labeled)}}\\
  &&&&Fungus dataset&84 images&\cite{234}&&&\\
  \bottomrule
  \end{tabular}}%
  \label{table3}
  \end{table}%
  \normalsize
  
\section{Discussion and Future Trends}
\label{sec4}
\subsection{For Weakly Supervised Learning Paradigm}
The two main goals of WSI analysis using the weakly supervised learning paradigm are global slide classification, which aims to accurately predict the labels of each WSI, and positive patch localization, which aims to accurately predict the labels of each positive patch in the positive bags. Among above two tasks, the former can be used for rapid automatic diagnosis of clinical pathology slides, such as early clinical screening, and the latter can be used for precise localization of tumor cells, as well as interpretable analysis of clinical diagnosis by deep learning networks. Based on the diagnostic results obtained from the whole slides, pathologists are often more interested in the precise location of tumor cells, the cell morphology and other microstructures for further analysis and corroboration. On the other hand, pathologists also expect new knowledge from the diagnosis of the deep neural networks, such as discovering new pathological patterns and structures, etc. A few current algorithms can perform the task of global slide classification well, but the task of positive patch localization is another challenge for most algorithms. A primary reason is that the loss functions of most bag-based deep MIL algorithms are defined only at the bag-level, and although mechanisms such as attention (Ilse \emph{et al.}~\citeyear{176}) can be used to measure the contribution of each instance to the bag-level classification, the network does not have enough motivation to classify all instances accurately (Shi \emph{et al.}~\citeyear{175}, Qu \emph{et al.}~\citeyear{203}). On the other hand, instance-based methods and hybrid methods, although defining instance-level classifiers, usually face a high risk of errors in pseudo-labeling or key instance selection. Therefore, it is a new challenge for the weakly supervised learning paradigm to further improve the ability to classify instances while obtaining a better slide-level diagnosis.

Further, with the emergence of the methods of the weakly supervised segmentation in the natural image processing field (Ru \emph{et al.}~\citeyear{179}, Xu \emph{et al.}~\citeyear{180}, Pan \emph{et al.}~\citeyear{181}, Lee \emph{et al.}~\citeyear{182}, Chen \emph{et al.}~\citeyear{183}), a new challenging direction for WSI analysis is to perform pixel-level semantic segmentation of the entire WSI based on weak or sparse labels. The task of the positive patch localization, which described in the previous section is still based on the classification of patches, and it is a more challenging task to further obtain pixel-level segmentation results based on the weak labels. A few current studies (Xu \emph{et al.}~\citeyear{178}, Qu \emph{et al.}~\citeyear{184}, Belharbi \emph{et al.}~\citeyear{185}, Lerousseau \emph{et al.}~\citeyear{186}) have made attempts in this new direction, but they still face many problems such as lack of details and precision on the segmentation results. Overall, for the weakly supervised learning paradigm, how to obtain the most detailed segmentation results as possible with weak labels is another promising study direction.

Another urgent need is the publicly available WSI datasets with fine-grained annotations at the patch level. As we all know, the scarcity of the publicly available pathological image datasets is an important factor hindering the development of the field. In recent years, we are grateful for the support of large public pathology datasets such as TCGA (\citeyear{160}), but public pathology datasets with fine-grained annotations are still in short supply for deeper research. To our knowledge, the large public WSI dataset with detailed annotation at the patch level is merely CAMELYON (Bejnordi \emph{et al.}~\citeyear{44}). We should encourage an individual or organization to provide more public WSI datasets with detailed patch-level annotations to promote the development of this study field.

\subsection{For Semi-Supervised Learning Paradigm}
For semi-supervised learning paradigm, a new study direction is the combination with active learning, the purpose of which is to use the most effective labeled data to obtain the highest performance. Active learning aims to find the most valuable samples in the unlabeled dataset to be annotated through iterative interactions with experts, which allows to further exploit the effects of semi-supervised learning. There are already a lot of studies on pathological image analysis with the help of active learning (Zheng \emph{et al.}~\citeyear{171}, Yang \emph{et al.}~\citeyear{41}) or combination with semi-supervised learning and active learning (Su \emph{et al.}~\citeyear{89}, Parag \emph{et al.}~\citeyear{188}).

Another challenge is the effect that noisy data and domain variation have on the performance of semi-supervised learning algorithms. In the field of computational pathology, noisy annotations are very common, because the instance features of pathological images are very complex and variable, and their sizes are so huge that doctors are likely to suffer from missing and mislabeling during annotation. When performing multicenter validation, significant staining variation between the slides from different centers is also very common as there is no uniform standard for staining pathological images among different centers. Both the noisy labels and the domain variation are powerful factors that affect the performance of semi-supervised learning in real-world scenarios. Recent studies (Koohbanani \emph{et al.}~\citeyear{125}, Cheng \emph{et al.}~\citeyear{74}, Shi \emph{et al.}~\citeyear{90}, Foucart \emph{et al.}~\citeyear{117}, Marini \emph{et al.}~\citeyear{69}) have made efforts on these two problems, and more studies in this field are expected.

\subsection{For Self-Supervised Learning Paradigm}
For self-supervised learning paradigm, although current relevant studies in the field of natural images are developing rapidly, the direct applications of these methods to pathological images will be hindered by the strong domain discrepancy (Ciga \emph{et al.}~\citeyear{157}, Koohbanani \emph{et al.}~\citeyear{125}). Therefore, how to design more effective self-supervised auxiliary tasks for pathological images is a promising direction.

On the other hand, self-supervised learning has been promoting the development of weakly supervised learning and semi-supervised learning in pathological image analysis. As we all know, it is difficult for a network to learn effective feature representations with very limited annotations. In contrast, self-supervised learning is very suitable for learning effective feature representations from a lot of unlabeled data. Therefore, it will be a popular way to combine the features extracted by self-supervised pre-training with the weakly supervised or semi-supervised downstream tasks in the future. On the one hand, the efficient feature representations obtained from self-supervised pre-training will greatly improve the efficiency of weakly supervised learning and semi-supervised learning, and on the other hand, weakly supervised learning or semi-supervised learning will fully release the new potential of self-supervised learning in the field of computational pathology.

\subsection{Limitations}
This review also has several limitations. First, due to space limitations, this review does not include more clinical studies. We focus more on top technical conferences and journals and do not include more excellent papers published in clinical journals. For more systematic reviews of clinical studies, see (Cifci \emph{et al.}~\citeyear{201}) and (Kleppe \emph{et al.}~\citeyear{202}) for details. In addition, since there are so many technical studies on artificial intelligence applied to computational pathology, it is difficult to summarize them all, and due to space limitations, we have tried to include as many recent articles as possible, while some of them have not been included.

\section{Conclusion}
\label{sec5}
In this review, we provide a systematic summary of recent studies on weakly supervised learning, semi-supervised learning, and self-supervised learning in the field of computational pathology from the theoretical and methodological perspectives. On this basis, we also present targeted solutions to some current difficulties and shortcomings in this field, and illustrate its future trends. Through a survey of over 130 papers, we find that the field of computational pathology is marching at high speed into a new era, which is automatic diagnosis and analysis with fewer annotation needs, wider application scope, and higher prediction accuracy.

\section*{Acknowledgments}
This work was supported by National Natural Science Foundation of China under Grant 82072021.

\scriptsize
\bibliography{survey}

\end{document}